\documentclass[10pt]{paper}

\usepackage{fullpage}
\usepackage{amsmath}
\usepackage{amsfonts}
\usepackage{amsthm}
\usepackage{hyperref}
\usepackage[capitalize]{cleveref}
\usepackage{array}
\usepackage{longtable}
\usepackage{multirow}
\usepackage[sort,numbers]{natbib}
\usepackage{float}
\usepackage{setspace}
\usepackage[dvipsnames]{xcolor}
\usepackage{tikz}
\usepackage{makecell}
\usepackage[export]{adjustbox}
\usepackage{listings, multicol}
\usepackage{xurl}

\lstdefinestyle{simplestyle}{
    commentstyle=\color{ForestGreen},
    keywordstyle=\color{RoyalBlue},
    numberstyle=\tiny\color{Gray},
    stringstyle=\color{Black},
    basicstyle=\small\ttfamily,
    emph={load,store},
    emphstyle=\color{Red},
    breakatwhitespace=false,         
    breaklines=true,                 
    captionpos=b,                    
    keepspaces=true,                 
    tabsize=3,
    language=Python, 
    belowskip=-2pt,
    morekeywords={forall,lock},
    deletekeywords=[2]{apply}
}

\theoremstyle{definition}
\newtheorem{subrule}{Rule}
\newcommand{\addrule}[2]{\begin{subrule}\label{#2}\textbf{#1}\end{subrule}}

\makeatletter
\g@addto@macro\@floatboxreset\centering
\makeatother

\makeatletter
\newcommand{\removelatexerror}{\let\@latex@error\@gobble}
\makeatother

\usetikzlibrary{positioning}
\usetikzlibrary {arrows.meta}
\usetikzlibrary {shapes.geometric}
\usetikzlibrary{fit}
\usetikzlibrary{shapes.symbols}

\newcommand{\onest}{\textsf{1}^{\!\textsf{T}}}

\newenvironment{nscenter}
 {\parskip=6pt\par\nopagebreak\centering}
 {\par\noindent\ignorespacesafterend}

\newcommand{\centergraphics}[1]{\begin{nscenter}\includegraphics[valign=t]{#1}\end{nscenter}}

\title{Blockbuster, Part 1: Block-level AI Operator Fusion}
\author{Ofer Dekel, Microsoft\\April, 2025}
\begin{document}
\twocolumn[\maketitle
\begin{abstract}
\emph{Blockbuster} is a framework for AI operator fusion in inference programs. The Blockbuster framework is compatible with any multiprocessor architecture that has a tiered memory hierarchy, including GPUs, multi-core CPUs, and some AI accelerator chips. It includes a graph-based representation for AI workloads, called a \emph{block program}, which explicitly models how blocks of data move between the memory tiers. It also includes an operator fusion procedure, which is made up of a candidate selection algorithm and a fusion algorithm that fuses each individual candidate -- this two-algorithm structure makes Blockbuster especially suitable for large AI programs. The current paper focuses on the fusion algorithm, which is a rule-based technique. While the literature is full of previous rule-based fusion algorithms, what sets our algorithm apart is its direct modeling of data movement between memory tiers, resulting in uniquely powerful fusion results. As a first sanity check, we demonstrate how our algorithm automatically rediscovers the well-known Flash Attention kernel. Then, we demonstrate the real power of our approach by fusing LayerNorm with matrix multiplication and RMSNorm with FNN-SwiGLU -- the latter involves fusing three matrix multiplications, a Hadamard product, a reduction, and a few elementwise operations into a single mega-kernel.  
\end{abstract}
\bigskip
]

\section{Introduction}
An AI graph compiler takes a high-level description of an AI program and generates an optimized low-level implementation of that program for a specific target computer, typically a powerful GPU or a dedicated AI accelerator chip. The AI program can represent either an inference or training workload, but the focus of this paper is on inference programs. This high-level description is often represented as a directed acyclic graph, where each node represents an \emph{operator} and each directed edge $U \rightarrow V$ represents an \emph{intermediate result} that is produced by $U$ and consumed by $V$. Some of the AI operators in the graph are \emph{standard operators}, taken from a predefined vocabulary of operators, such as the ones defined in PyTorch \citep{PyTorchOps}, TensorFlow \citep{TFOps}, or ONNX \citep{ONNXOps}. Other operators are \emph{custom operators}, which are bespoke functions, written for a particular program. Concrete examples of graph representations of AI programs include TensorFlow Computation Graphs \citep{Abadi2015}, Torch FX graphs \citep{Suo2020}, and ONNX graphs \citep{Kuchaiev2019}.    

One of the key optimization techniques used by AI graph compilers is \emph{operator fusion} (a.k.a. \emph{kernel fusion}). This technique combines adjacent operators into bigger \emph{fused operators}, reducing unnecessary data movement and kernel invocation overhead. In this paper, we present a new framework for AI operator fusion named \emph{Blockbuster}, which includes a new representation for AI programs and a fusion algorithm that operates on that representation. 

The Blockbuster framework is compatible with any multiprocessor computer that has at least two tiers of memory: each of its processors has a small-and-fast \emph{local memory} and all of them share a large-but-slow \emph{global memory} (an extension to more complex memory topologies is straightforward). Data can be copied back-and-forth between the global memory and each of the local memories, but it cannot be copied directly from one local memory to another. The inputs of the program start in global memory, and the output of the program must end up in global memory, but each processor can only operate on data that is located in its local memory. This abstract machine model fits the general design of most GPUs (processors map to GPU streaming-multiprocessors, local memory maps to GPU shared memory), multi-core CPUs (processors map to CPU cores, local memory maps to each core's cache), and some AI accelerator chips. 

The inputs, outputs, and intermediate results of an AI program are typically large multidimensional arrays. Each operator in the program takes one or more arrays as input and generates one or more arrays as output. Therefore, we refer to the AI program as an \emph{array program} (a.k.a. \emph{tensor program}) and to it operators as \emph{array operators}. We assume that these large arrays fit in global memory but are too big to fit in local memory. To overcome this limitation, each array is split into blocks, which are small enough to fit in local memory. Individual blocks are incrementally loaded into local memory, the corresponding processor performs various operations on them, and the results of those operations are copied back to global memory. 

If the compiler doesn't carefully schedule how blocks are copied between the global and local memories, the processors may end up waiting idly for their input data. Operator fusion decreases the overhead of copying blocks between global and local memory by combining multiple operators into bigger fused operators. If two subsequent operators $U \rightarrow V$ are fused into a single operator, the intermediate result produced by $U$ and consumed by $V$ may not need to be materialized in the global memory. Similarly, if a block from an input array $A$ is consumed by two operators, $U$ and $V$, fusing these operators may allow the blocks of $A$ to be copied once from global to local memory instead of twice. If each kernel launch incurs a fixed cost, reducing the number of operators in the program has the additional advantage of reducing the total kernel launch overhead.  

We illustrate the idea behind operator fusion with a simple and well-known example. Let $A$ and $B$ be two large matrices stored in global memory. Assume that $A$ is split into row-wise blocks and $A[i]$ represents its $i$'th row-block, the input matrix $B$ is split into column-wise blocks and $B[j]$ represents its $j$'th column-block. The sizes of these blocks are chosen according to the capacity of the local memory. The goal is to multiply $A$ and $B$, and then to apply the elementwise \emph{RELU} activation function to their product. Naively, we would execute the two operations separately: load each pair of blocks, $A[i]$ and $B[j]$, into local memory, multiply them, and store the intermediate result in global memory; once the entire matrix multiplication operation has completed, load the intermediate result block-by-block into local memory to apply the RELU function. This naive implementation is represented by the following code:

\begin{lstlisting}[style=simplestyle]
for m in range(M):
	t1 = load(A[m])
	for n in range(N):
		t2 = load(B[n])       
		t3 = t1@t2 
		store(t3, I[m,n])        

for m in range(M):
	for n in range(N):
		t4 = load(I[m,n])
		t5 = RELU(t4)
		store(t5, C[m,n])
\end{lstlisting}

Note that we use lower case variables, such as \texttt{t1}, to represent temporary values in local memory, and upper case variables, like \texttt{I}, to represent intermediate results that are materialized in global memory. We  use \texttt{load} and \texttt{store} instructions to explicitly highlight when data is loaded into local memory and stored back in global memory. 

Instead of performing multiplication and RELU separately, we can fuse these two operations into a single kernel and avoid materializing the large intermediate array \texttt{I} in global memory. The fused implementation is represented by the following code:

\begin{lstlisting}[style=simplestyle]
for m in range(M):
	t1 = load(A[m])
	for n in range(N):
		t2 = load(B[n])       
		t3 = t1@t2
		t4 = RELU(t3) 
		store(t4, C[m,n])        
\end{lstlisting}

Note that the intermediate result \texttt{t3} is never materialized in global memory, and is directly processed by the subsequent RELU operator. The goal of an operator fusion algorithm is to automatically discover these kinds of fusion opportunities in any AI program. 

The first part of the Blockbuster framework is a new graph-based representation called a \emph{block program}, presented in \cref{sec:rep}, which explicitly models how blocks move between global and local memories. The nodes of a block program include input nodes, output nodes, and \emph{block operator} nodes, which are similar to array operators, except that they operate on blocks instead of full arrays. Each array operator from the original array program maps to a predefined subgraph in the block program. Modeling the AI workload at the block level enables our fusion algorithm to explicitly optimize how data blocks move between the memory tiers when the program is executed. 

The second part of the Blockbuster framework is a new rule-based fusion algorithm. We start by defining a set of logic-preserving substitution rules in \cref{sec:sub} -- each rule attempts to find a local subgraph in the block program that matches a certain pattern and replaces that subgraph with a logically-equivalent substitute. Some of these rules directly exploit fusion opportunities in the block program and reduce the amount of data copied between global and local memories. Other rules are designed to reveal hidden patterns in the block program that can be exploited by a subsequent substitution rule. Some of the rules replicate work and increase the computational load of the program in exchange for exposing hidden fusion opportunities. In \cref{sec:algo} we describe our fusion algorithm, which merely applies the substitution rules in a particular order. 
 
Blockbuster also includes a third part, which is a fusion-candidate selection algorithm. That algorithm is not covered in this paper and is instead deferred to an upcoming companion paper \citep{Dekel2025b}. Fusion candidates are subgraphs of the block program that are suitable for fusion. Fusion-candidate selection is largely independent from the specific fusion algorithm presented in this paper, and we can discuss the two algorithms separately as long as we precisely define the contract between them. In broad strokes, the selection algorithm identifies a set of fusion candidates and the fusion algorithm attempts to fuse each candidate separately. After the fusion algorithm is done, the selection algorithm evaluates the fused results and chooses the optimal set of fused kernels that implements the entire block program. In fact, the selection algorithm can even send candidates to several different fusion algorithms, compare their results, and choose the best ones. Subgraphs of the original block program are themselves block programs, so the fusion algorithm doesn't need to be aware that it is operating on a subset of a larger program. The selection algorithm is also responsible for choosing the block shapes, because those need to be consistent across the different kernels chosen to implement the program.

Although the fusion algorithm does not need to know precisely how the selection algorithm chooses its candidates, the existence of a selection algorithm allows us to make certain simplifying assumptions about the fusion problem. First, we can assume that our fusion algorithm only receives fusion candidates that are entirely made up of standard operators. If we want to also fuse custom operators, the selection algorithm can send those candidates to a different fusion algorithm, which is designed for fusing custom operators.

Second, our fusion algorithm doesn't have to worry about the risks of \emph{excessive fusion}. Excessive fusion can occur when we aggressively fuse very large operator graphs, instead of breaking them up into smaller pieces and fusing each piece separately. Excessively fused programs increase the pressure on local memory and run the risk of exceeding the local memory. This can be overcome by choosing a smaller block size, but smaller blocks can sometimes hurt performance. Overall, fusing the largest possible subgraph doesn't always lead to the best performance. However, our fusion algorithm can ignore the dangers of excessive fusion because the selection algorithm makes sure to choose fusion candidates of the right size. In fact, the Blockbuster selection algorithm does this in a provably optimal way \citep{Dekel2025b}. Not having to worry about excessive fusion makes our framework especially suitable for large programs, such as an entire Decoder block in the Transformer architecture \citep{Vaswani2017}.  

Third, since the selection algorithm knows how to evaluate fused candidates and choose an optimal subset of kernels, it implies that the fusion algorithm can produce multiple different fused versions of each candidate and the selection algorithm will know how to pick the best one. This will come in handy when our fusion algorithm considers taking risky steps that could backfire, such as replicating work in order to fuse more aggressively.

As noted above, it is the selection algorithm's responsibility to set the block shapes. Choosing ideal block shapes is critical for achieving peak performance. A convenient feature of our fusion algorithm is that its choices do not depend on the block shapes. This means that the selection algorithm can invoke the fusion algorithm once per candidate, without specifying the block shapes in advance, and then optimize all the shapes after-the-fact. 

After presenting the block program representation, the substitution rules, and the algorithm that applies them, we present a three step-by-step examples of our fusion framework in action, in \cref{sec:examples}. For begin each example by translating an array program into our block program representation. Then, we apply our fusion algorithm and show the result of each step. The final outcome of each example is a single fully-fused kernel, which does not store any intermediate results in global memory. 

The first example demonstrates how our algorithm automatically rediscovers the well-known \emph{Flash Attention} kernel \citep{Dao2022, Dao2023, Dao2024}, which is one of the most compelling examples of operator fusion. Flash Attention is a fused implementation of the \emph{Attention} operator, which includes a matrix multiplication, followed by an elementwise division-by-constant, followed by a softmax operation, followed by another matrix multiplication. Note that Flash Attention also applies a numerical stabilization technique, sometimes referred to as \emph{online softmax}, which is independent from operator fusion and can be applied separately after fusion. Therefore, we ignore the numerical stabilization technique in the body of this paper, focus exclusively on the fusion part, and then, for completeness, outline the numerical stabilization technique in an Appendix. 

Flash Attention had to be manually discovered by humans because state-of-the-art operator fusion algorithms where not sophisticated enough to discover it automatically. The algorithm presented in this paper automatically rediscovers Flash Attention when given the block program for straightforward (unoptimized) Attention. To the best of our knowledge, \citet{Wu2024} is the only other fusion approach capable of rediscovering Flash Attention. Note that if the straightforward definition of Attention is hidden within a much larger program, our selection algorithm will identify it and select it as a candidate. 

Finally, we present two more examples of useful array programs that are successfully fused by our algorithm. The fused operators that result from these examples are of independent value and interest. The first one is LayerNorm fused with matrix multiplication -- one could call the resulting fused kernel \emph{Flash-LayerNorm+Matmul}. The second one is RMSNorm fused with an entire FFN-SwiGLU subgraph, which is a common pattern that appears in Llama, DeepSeek, and other modern LLM architectures, and involves three matrix multiplications, a Hadamard product (elementwise matrix multiplication), a reduction, and a few elementwise operations. For lack of a better name, one could call the resulting fused mega-kernel \emph{Flash-RMSNorm+FFN-SwiGLU}. We are not aware of any previous operator fusion framework that can produce these advanced kernels. 

In summary, this work introduces the \emph{Blockbuster} framework for AI operator fusion and makes the following contributions:
\begin{itemize}
    \item The block program representation for AI workloads, which explicitly represents how blocks of data
    move between memory tiers and captures the information needed by fusion algorithms. 
    \item A set of block program substitution rules and a simple rule-based fusion algorithm that applies these rules. 
    \item \sloppy{The Flash-LayerNorm+Matmul and Flash-RMSNorm+FFN-SwiGLU kernels,} which are automatically discovered by our algorithm. 
\end{itemize}

\subsection*{Related Work}

There is an extensive bibliography of past work on operator fusion. Most of this previous work is more sophisticated than our simple rule-based approach, but to the best of our knowledge, none of these fancier techniques can produce results similar to ours. In this section, we give a few examples of related work and compare it with our own approach. 

Many of the existing fusion algorithms are rule-based. While our algorithm is a straightforward rule-based approach, many of the previous algorithms enhance the rule-based paradigm with clever refinements. For example, \citet{Jia2019} generates substitution rules automatically and \citet{Niu2021} includes an extensive set of rules based on mathematical properties, including associative, distributive, and commutative laws for different algebraic operations. In contrast, we manually define a small set of substitution rules and apply them in a specific priority order.  

Some fusion techniques emphasize the importance of using a carefully designed cost function to guide the search. For example, \citet{Zheng2020c} uses a coarse cost function to quickly screen the initial candidates, and then switches to a more refined cost function for the final code generation phase. Other techniques focus on the strategy used to explore the search space. For example, \citet{Zheng2020} draws samples from a hierarchical representation of the search space and then uses uses those samples to seed an evolutionary search strategy. \citet{Jia2019} and \citet{Wu2024} incorporate pruning techniques to reduce the vast size of their search spaces. Our approach does not explicitly define a cost function or a systematic way of exploring the search space, but merely applies a set of substitution rules in a specific order, and leaves the rest to a global candidate-selection algorithm. 

Another line of research relies on powerful optimization techniques, including numerical optimization algorithms and theorem provers, to guide the search and to verify its correctness. For example, \citet{Jia2019}, mentioned above, uses a theorem prover to ensure the correctness of their automatically-generated rules. \citet{Zheng2020c} formulates the selection of optimal fusion plans as an Integer Linear Program and \citet{Huang2021} applies operator fusion alongside other optimization techniques using Mixed Integer Programming. \citet{Wu2024} uses a probabilistic equivalence verification procedure to ensure that the optimized kernel retains the correct logic. Our approach does not require any correctness verification, since all we do is apply a sequence of logic-preserving rules. 

A common theme in several fusion algorithms is to categorize the different AI operators or the interactions between them into categories, and then define fusion strategies for each category. \citet{Niu2021} categorizes operators into the categories One-to-One, One-to-Many, Many-to-Many, Reorganize, and Shuffle. \citet{Hu2024} performs a fission step that breaks complex AI operators into more basic operators: Elementwise, Reduce, Broadcast, LayoutTransformation, and Linear. Performing fission has the benefit of enabling the fusion algorithm to fuse parts of one operator with parts of another. Our block program representation resembles a post-fission graph and the Map, Reduce, and Elementwise operators in our block programs resemble some of the categories mentioned above. 

Some previous approaches to operator fusion also split the problem into two separate tasks: candidate selection and candidate fusion. \citet{Zhao2022} calls this technique Partition-Based Operator Fusion, first using aggregation rules to piece together fusion candidates from individual operators and then using Polyhedral Loop Optimization and other fusion techniques on each candidate. \citet{Cai2023} presents a selection algorithm based on dynamic programming followed by a simple fusion algorithm that invokes each operator as soon as its first input block is ready. The Blockbuster framework borrows from these ideas and splits the work between a selection algorithm and a fusion algorithm.

In summary, many of the ideas used in our Blockbuster framework already appear in previous algorithms. Moreover, our approach may seem quite simplistic compared to some of the more sophisticated previous art. Nevertheless, our work is noteworthy due to the surprisingly powerful bottom-line results that we achieve, which we haven't seen in the existing literature. 

\section{Block Programs}
\label{sec:rep}
The original array program is a directed acyclic graph of operators that processes large multidimensional arrays. To perform fusion effectively, we require a more detailed representation that explicitly describes how these large arrays are split into smaller blocks, and how those blocks move between the global and local memories. To that end, we introduce the block program representation.

\subsection{Block Program Graph Structure}
The block program is a hierarchical directed-acyclic-graph, which means that some of its nodes contain block program graphs inside of them. The block program's top-level graph is logically equivalent to the original array program and its inner graphs correspond to sections of the original program. 

The block program has the same inputs and outputs as the original array program, which are represented by input nodes and output nodes respectively. For simplicity, assume that the inputs and outputs are all matrices and note that the generalization to higher dimensions is straightforward. The input and output matrices are typically too big to fit in local memory, and are therefore split into blocks along both dimensions. The number of blocks along each dimension is a parameter, which can later be optimized using an auto-tuning procedure. Note that we could always set the number of blocks along either dimension to $1$, which would be equivalent to splitting the matrix along only one of its dimensions. 

Each row of matrix blocks is stored in global memory as a list of blocks, and the entire matrix is therefore stored as a list of lists-of-blocks. That is, we assume that the inputs of a block program are row-major -- if they happen not to be row-major, we must reorder and transpose the blocks as they are copied into local memory. 

Assume that the blocks are small enough so that a few blocks from different matrices can fit comfortably in local memory, and assume that the block shapes are compatible with each other. For example, if the program sums two blocks, their shapes must be identical; if it multiplies two blocks, the number of columns in the left-hand block must equal the number of rows in the right-hand block; etc. The compatibility between block shapes is defined more precisely later on, when we introduce functional operators.    

An edge $U \rightarrow V$ in the block program represents an intermediate result, produced by $U$ and consumed by $V$, just like an edge in an array program. Additionally, this edge also implies a strict precedence constraint: operator $V$ cannot begin any work before operator $U$ completes all of its work. Each edge is either \emph{unbuffered} or \emph{buffered}. An unbuffered edge $U \rightarrow V$ represents a single block, a vector, or a scalar: $U$ produces the intermediate result in local memory and $V$ consumes it directly from there, without going through a global memory buffer. Since each processor has its own local memory, an unbuffered edge $U \rightarrow V$ implies that the operators $U$ and $V$ are executed on the same processor. A buffered edge represents an intermediate result that doesn't fit in local memory, namely, a list of blocks, list of vectors, or a list of lists. Since $U$ must produce the entire intermediate result before $V$ can begin, it stores the result in a global memory buffer. After $U$ terminates, $V$ can start incrementally consuming the data from this buffer. Edges that are incident with the inputs and outputs of a block program are always buffered edges, since we assumed that the inputs and outputs reside in global memory. We highlight buffered edges in our diagrams by coloring them red. 

Recall that the primary goal of fusion is to avoid materializing large intermediate results in global memory and to minimize the number of passes over input data -- this goal precisely corresponds to removing buffered edges from a block program.

Most of the nodes in a block program represents block operators, which come in four types: \emph{functional operators} represent functions on individual blocks, vectors, or scalars, \emph{map operators} represent embarrassingly parallelizable list-to-list operations, \emph{reduction operators} represent reduction operations, which summarize a list into a single item, and \emph{miscellaneous operators} represent everything else. 

\paragraph{Functional operators}\!\!\!represent functions (stateless operations) whose inputs and outputs are blocks, vectors, or scalars in local memory. Each functional operator has a mathematical description and a set of constraints on its inputs and outputs. Recall that we previously assumed that the block shapes are compatible with each other; equivalently assume that the block shapes satisfy all the functional operator constraints in the block program.     

\begin{table*}[t]
\caption{Examples of functional operator definitions}
\label{tab:opnodes}
\begin{tabular}{| p{.16\textwidth} | >{\raggedright}p{.21\textwidth} | p{.30\textwidth} | p{.21\textwidth} |}
    \hline
    \textbf{Operator} & \textbf{Description} & \textbf{Constraints} & \textbf{Numpy Definition} \\
    \hline
    \texttt{r = add(a,b)} & block addition & \texttt{a.shape==b.shape and a.shape==r.shape} & \texttt{r = a+b} \\
    \hline
    \texttt{r = mul(a,b)} & block elementwise multiplication & \texttt{a.shape==b.shape and a.shape==r.shape} & \texttt{r = a*b} \\
    \hline
    \texttt{r = row\_shift(a,c)} & add a value to each row of a block& \texttt{a.ndim==2 and c.ndim==1 and a.shape[0]==c.size and a.shape==r.shape} & \texttt{r = a+c[:,newaxis]} \\
    \hline
    \texttt{r = row\_scale(a,c)} & multiply each row of a block by the corresponding value& \texttt{a.ndim==2 and c.ndim==1 and a.shape[0]==c.size and a.shape==r.shape} & \texttt{r = a*c[:,newaxis]} \\
    \hline
    \texttt{r = row\_sum(a)} & sums the values in each row of a block & \texttt{a.ndim==2 and r.ndim=1 and a.shape[1]==r.size} & \texttt{r = sum(a,axis=0)} \\
    \hline
    \texttt{r = dot(a,b)} & multiply a block with the transpose of another block & \texttt{a.ndim==2 and b.ndim==2 and a.shape[1]==b.shape[1] and (a.shape[0],b.shape[0])== r.shape} & \texttt{r = a@(b.T)} \\
    \hline  
    \texttt{r = outer(a,b)} & outer-product of two vectors & \texttt{a.ndim==1 and b.ndim==1 and (a.size,b.size)== r.shape} & \texttt{r = outer(a,b)} \\
    \hline  
\end{tabular}
\end{table*}

In particular, functional operators can represent elementwise operations on blocks. An elementwise operation is any scalar function, which is applied independently to each element of a block or vector. For example, imagine the simple block program that takes the input block \texttt{a}, applies the operation $x \mapsto (x-s)/d$ (where $s$ and $d$ are constants) to each element, and produces the output block \texttt{b}. This block program is represented by the graph:
\centergraphics{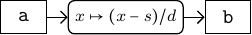}

The constraint associated with an elementwise operation is that the input and output blocks have the same shape. Since \texttt{a} and \texttt{b} are individual blocks, the edges of this graph are unbuffered edges.

Functional operators can also represent many other operations that are not elementwise, with useful examples presented in \cref{tab:opnodes}. 

Functional operators can be used to create block programs with multiple inputs and outputs. 
\centergraphics{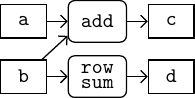}

For example, the program above takes two blocks, \texttt{a} and \texttt{b}, adds them together (the output is a block \texttt{c}) and also calculates a row-wise sum of the second block (the output is a column vector \texttt{d}).
\paragraph{Map operators}\!\!\!\!contain an inner block program graph and behave like an embarrassingly-parallelizable for-loop over the inner program. In other words, a map iterates over one or more lists (of equal length) and applies the inner graph independently on each iteration. The output of the map operator is one or more output lists, of the same length as the input lists. Besides the input lists, the map operator can also take individual blocks, vectors, or scalars as additional inputs.

For example, imagine that the input is a list of blocks and the goal is to apply the elementwise operation $x \mapsto (x-s)/d$ to each element in each block. This block program is represented by the following graph:
\centergraphics{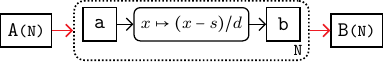}
The input $A$ and output $B$ are both lists of $N$ blocks, as indicated by the notation $A(N)$ and $B(N)$. The map operator is drawn with a dotted line and the iteration dimension is identified by the $N$ in its bottom-right corner. The inner graph begins with a block $a$ in local memory and ends with a block $b$ in local memory. Since $A$ resides in global memory, the input node's outgoing edge is a buffered edge, which appears in red. This block program can also be represented as code:

\begin{lstlisting}[style=simplestyle]
forall n in range(N):
	a = load(A[n])
	b = (a-s)/d
	store(b,B[n])
\end{lstlisting}

The \texttt{forall} instruction in this code represents an embarrassingly parallelizable for-loop. For brevity, whenever the input and output of the inner graph are obvious from the context, we omit them from the diagram: 
\centergraphics{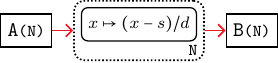}

Next, consider the case where the input and output matrices are blocked along two dimensions, which means that they are stored in global memory as lists of lists-of-blocks. To apply the elementwise operation to each block, we nest a map operator inside another map operator, as follows:
\centergraphics{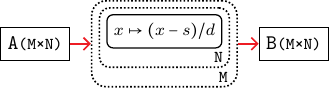}

We also introduce a shorthand for nested map operators, where the iteration dimensions appear in a comma-separated list (outer loop first):
\centergraphics{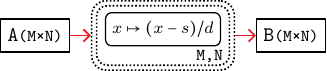}

The corresponding code for this block program is:
\begin{lstlisting}[style=simplestyle]
forall m in range(M):
	forall n in range(N):
		a = load(A[m,n])
		b = (a-s)/d
		store(b,B[m,n])
\end{lstlisting}
For an example of a map operator that takes multiple input lists and produces multiple output lists, recall the example given above of a block program that adds two blocks and also sums the columns of one of them. Now imagine that this program is applied $N$ times in a parallelizable loop. The resulting block program becomes:
\centergraphics{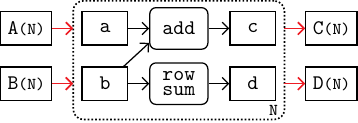}

This graph is an example of a case where we can't simply omit the input and output nodes from the inner graph, because their position in the graph plays a nontrivial role. Instead, we introduce a more concise look, where we replace the inner graph's input and output nodes with small circles on the perimeter of the map operator:
\centergraphics{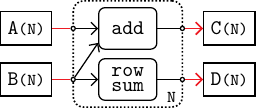}
We still prefer to completely omit input and output nodes from inner graphs whenever their role is obvious.

\paragraph{Reduction operators}\!\!\!\!represent reduction operations on lists. Their input is a list of items, such as a list of blocks, and their output is a single-item summary of the list. The most common reduction operation is addition, which sums all the list items together. 

For example, imagine that the goal is to sum each row of a matrix that is split horizontally into blocks. We first use a map to independently apply the \texttt{rowsum} operator (described in \cref{tab:opnodes}) to each block. The output of this map is a list of column vectors, which contain the partial sums of the corresponding blocks. Then, we reduce these partial sums into a single vector. The block program graph looks like this:
\centergraphics{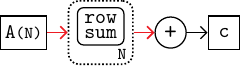}
The addition reduction operator is represented by a circled plus sign. Note that the output of the map operator requires a buffered edge while the single vector generated by the reduction uses an unbuffered edge. 

If the input matrix in the previous example were split along both dimensions, it would be stored as a list of lists-of-blocks. We could handle this situation by introducing a map operator that applies the program described above to each list-of-blocks. The result would be:
\centergraphics{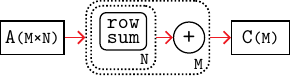}

Additional examples of block programs that include reduction operators are given in \cref{tab:subgraphs}.

\paragraph{Miscellaneous operators}\!\!\!\!are used as a last resort, to represent array operators that can't be represented using the previous operator types. For example, an array operator that sorts the elements of each row in a large matrix can't be expressed using the previous block operator types, and therefore becomes a miscellaneous operator in the block program. Custom operators in the original array program also become miscellaneous operators in the block program. The existence of miscellaneous operators simply ensures that any array program can be converted into a block program. 

\subsection{Operator Representations}
\begin{table*}[t]
\caption{Examples of array program operators as represented in the block program.}
\label{tab:subgraphs}
\begin{tabular}{| p{.28\textwidth} | p{.65\textwidth} |}
    \hline
    \textbf{Array Program Operator} & \textbf{Block Program Subgraph} \\
    \hline
    elementwise function $f(x)$& 
    \trimbox{0 -6pt 0 -6pt}{\includegraphics[valign=c]{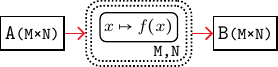}}\\
    \hline
    \texttt{matmul} & 
    \trimbox{0 -6pt 0 -6pt}{\includegraphics[valign=c]{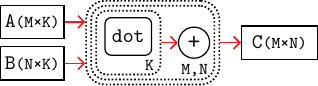}}\\
    \hline
    \texttt{softmax} & 
    \trimbox{0 -6pt 0 -6pt}{\includegraphics[valign=c]{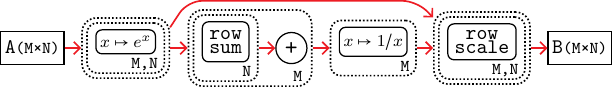}} \\ 
    \hline
    \texttt{rmsnorm} & 
    \trimbox{0 -6pt 0 -6pt}{\includegraphics[valign=c]{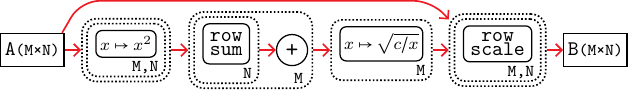}}    
    \\
    \hline
\end{tabular}
\end{table*}

Given an array program as input, the first step is to convert it into a block program. This conversion is done by a straightforward lookup: each operator in the array program is replaced with a predefined subgraph in the block program. If an operator from the original program does not appear in the lookup table, it becomes a miscellaneous operator in the block program. The predefined subgraphs for several operators are presented in \cref{tab:subgraphs} and adding other standard operators to the lookup table is left as an exercise for the reader. 

Note that the subgraphs in \cref{tab:subgraphs} are all fully unfused and use global memory extensively, even when a straightforward fusion opportunity is evident. Additionally, these subgraphs ignore numerical stability issues and assume that the computation is performed with infinite-precision arithmetic (we can add numerical stability after-the-fact, as described in the Appendix).

\section{Substitution Rules}
\label{sec:sub}
A substitution rule is a logic-preserving transformation that matches a small subgraph in the block program and replaces it with a different subgraph. The matched subgraph is called the \emph{pattern} and the replacement is called the \emph{substitution}. Substitution rules can be applied to the top-level graph of the block program or to any of the inner graphs that reside inside map operators. 

Each rule implements a \texttt{match} function, which takes a graph \texttt{G}, searches it for a subgraph that matches the rule's pattern, and returns all the necessary information about the match: 
\begin{lstlisting}[style=simplestyle]
M = rule1_match(G)
\end{lstlisting}
If multiple subgraphs match the pattern, the function chooses one of them arbitrarily, and if no matches are found, the function returns \texttt{None}. The match function may take optional arguments that constrain the search in different ways, for example, if a rule matches a map operator, it can be required to have a specific dimension:
\begin{lstlisting}[style=simplestyle]
M = rule1_match(G,dim="d")
\end{lstlisting}
Assume that \texttt{M} remembers the rule and the graph that created it, which allows us to define a global \texttt{apply} function that performs the substitution that corresponds to \texttt{M} and returns the modified graph:
\begin{lstlisting}[style=simplestyle]
G_new = apply(M)
\end{lstlisting}
Additionally, \texttt{M} can have attributes that describe different properties of the match. For example, if the pattern contains a map operator, the attribute \texttt{M.dim} names the dimension of map operator that was matched.

Our framework includes two types of substitution rules. \emph{Fusion rules} directly advance the goal of removing buffered edges, while \emph{companion rules} have the more indirect role of exposing hidden patterns in the graph, which can eventually be exploited by subsequent rules. Some companion rules are especially aggressive and will even replicate work to expose fusion opportunities. 

\subsection{Fusion Rules} \label{subsec:fusion}
Fusion rules directly fuse adjacent operators in the block program and have the immediate and obvious benefit of eliminating at least one buffered edge. 

\begin{minipage}{\columnwidth}
\addrule{Fuse Consecutive Maps}{rule:consecutive}
\begin{nscenter}
\begin{tabular}{c c c}
Pattern:&~~&Substitution:\\[8pt]
\includegraphics{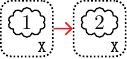}&&\includegraphics{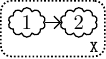}   
\end{tabular}
\end{nscenter}
\end{minipage}

Clouds in the figure above represent arbitrary inner graphs. The pattern for this rule is two consecutive map operators, $U \rightarrow V$ that share the same dimension. There may be an opportunity to fuse $U$ and $V$, but only if an important condition is met: there cannot be any indirect (multi-edge) path from $U$ to $V$. More precisely, if there happens to exist an indirect path between them -- from $U$ to a third operator $W$ and from there to $V$ -- fusing $U$ and $V$ would create an illegal loop in the graph: a path from the new fused operator to $W$ and back. On the other hand, if the only paths from $U$ to $V$ are direct unbuffered edges between them, we can safely fuse the maps.

Fusing $U$ and $V$ means replacing them with a single map and concatenating their inner graphs. The fused operator inherits the inputs and outputs from the original maps, except for the unbuffered edges between $U$ and $V$, which are replaced by corresponding edges in the new inner graph. 

\begin{minipage}{\columnwidth}
\addrule{Fuse Sibling Maps}{rule:sibling}
\begin{nscenter}
\begin{tabular}{c c c}
Pattern:&~~~~~&Substitution:\\[6pt]
\includegraphics[valign=c]{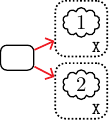}&&\includegraphics[valign=c]{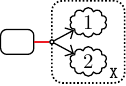}
\end{tabular}
\end{nscenter}
\end{minipage}

As before, clouds represent arbitrary inner graphs. The pattern for this rule is two map operators, with the same dimension, that share a common parent. Similar to \cref{rule:consecutive}, there may be an opportunity to fuse them if a condition is met: the two maps cannot be reachable from each other. If there exists an edge between the two maps, that case is already covered by \cref{rule:consecutive}, and if there exists an indirect path between the two maps, fusing them would create a loop. If the condition is met, we can safely fuse the maps.

The original inner graphs are combined into a single inner graph. The fused map inherits all the inputs and outputs from the original maps, except for the two incoming edges from the shared parent, which are merged into a single shared edge.

\begin{minipage}{\columnwidth}
\addrule{Fuse Map with Reduction}{rule:mapreduce}
\begin{nscenter}
\begin{tabular}{c c c}
Pattern:&~~&Substitution:\\[8pt]
\includegraphics{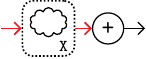}&&\includegraphics{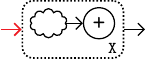}   
\end{tabular}
\end{nscenter}
\end{minipage}

If a map operator is followed by a reduction operator, they can be fused. Instead of performing the entire map, storing the intermediate result in a global memory buffer, and reading the result back to local memory to perform the reduction, we can calculate the reduction on-the-fly while executing the map. We depict this by moving the reduction operator inside the map operator and changing the map's output edge to an unbuffered edge. 
Consider the following example:
\centergraphics{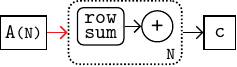}
The simplest way to implement the fused operation is to replace the parallel for-loop with a serial one: 

\begin{lstlisting}[style=simplestyle]
for n in range(N):
	a = load(A[n])
	b = row_sum(a)
	c += b
\end{lstlisting}
Another possibility is to use the commutative law for addition and apply an atomic addition operation:

\begin{lstlisting}[style=simplestyle]
forall n in range(N):
	a = load(A[n])
	b = row_sum(a)
	with lock:
		c += b
\end{lstlisting}

The code above uses a lock to define a critical section, inside which the addition is safely performed.

\subsection{Companion Rules} \label{subsec:companion}
Companion rules do not have an immediate benefit. Their goal is to expose fusion opportunities that can later be exploited by a fusion rule. 

\begin{minipage}{\columnwidth}
\addrule{Linearity of Matmul: Swap Scale/Dot}{rule:scale_dot}
\quad Pattern:
\centergraphics{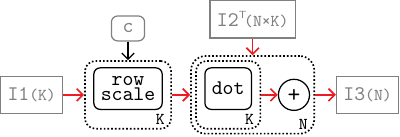}
\quad Substitution:
\centergraphics{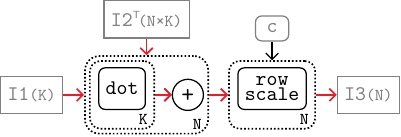}
\end{minipage} \\ 

This rule exploits the linearity of matrix multiplication. The pattern is a mapped \texttt{row\_scale} operation followed by a matrix-multiplication (a mapped dot-and-accumulate). The rule can only be applied if the mapped scale operator has no other outgoing edges, since the output of this operator will no longer be computed after the substitution. The rule swaps the order of the two operations, so that the scaling occurs after the multiplication. Mathematically, this rule is simply the associative law for matrix multiplication: 
$$
\textrm{diag}(c) \cdot I_1 \cdot I_2 ~=~ \textrm{diag}(c) \cdot (I_1 \cdot I_2)~~.
$$
Specifically, the pattern assumes that \texttt{I1} is a single row-of-$K$-blocks, and that $I2$ is a matrix of $K \times N$ blocks. This pattern occurs whenever the left-hand matrix undergoes a row-wise normalization before multiplication, such as in \texttt{LayerNorm}, \texttt{RMSNorm}, or \texttt{Softmax}. 

Swapping the order of the two maps sometimes creates the right circumstances for \cref{rule:consecutive} or \cref{rule:sibling} to kick in. Additionally, note that the dimension of the scaling map changes from $K$ in the pattern to $N$ in the substitution, which can also create opportunities for subsequent rules. Finally, note that the matrix multiplication in the pattern cannot be performed before the scaling vector \texttt{c} is ready, but the matrix multiplication in the substitution does not suffer from this bottleneck.  

The matched pattern can be represented with code:
\begin{lstlisting}[style=simplestyle]
forall k in range(K):
	t1 = load(I1[k])
	t2 = t1*c[:,newaxis]
	store(t2,I4[k])
forall n in range(N):
	t3 = 0
	for k in range(K):
		t4 = load(I4[k])
		t5 = load(I2[k,n])
		t3 += t4@t5	
	store(t3,I3[n])

\end{lstlisting}
Note that \texttt{c} is a column vector in local memory. The code for the substitution is:
\begin{lstlisting}[style=simplestyle]
forall n in range(N):
	t1 = 0
	for k in range(K):
		t2 = load(I1[k])
		t3 = load(I2[k,n])
		t1 += t2@t3	
	store(t1,I4[n])
forall n in range(N):
	t4 = load(I4[n])
	t5 = t4*c[:,newaxis]
	store(t5,I3[n])

\end{lstlisting}
\begin{minipage}{\columnwidth}
\addrule{Linearity of Matmul: Swap Shift/Dot}{rule:shift_dot}
\quad Pattern:
\centergraphics{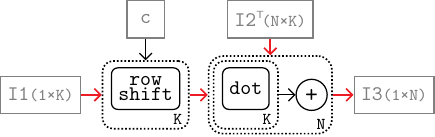}
\quad Substitution:
\centergraphics{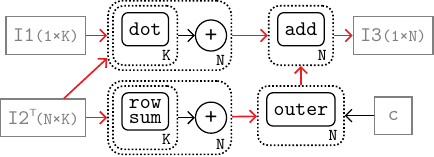}
\end{minipage} \\

This pattern is similar to the one in \cref{rule:scale_dot}, with an additive shift instead of a multiplicative scaling. In other words, the pattern is a mapped shift operation followed by a matrix multiplication. The rule can only be applied if the mapped shift operator has no other outgoing edges, since the output of this operator is no longer computed after the substitution. 

Adding a column vector $c$ to each column of a matrix $I_1$ is mathematically represented as 
$$
I_1 + c \cdot \onest ~~,
$$
where $\onest$ is the row-vector of ones, whose size equals the number of columns in $I_1$, and $c \cdot \onest$ represents the outer product. The substitution for this rule is simply the result of the distributive property of matrix multiplication:
$$
\left( I_1 + c \cdot \onest \right) \cdot I_2 ~=~ I_1 \cdot I_2 + c \cdot \left(\onest \!\cdot I_2 \right)~~.
$$
Note that $\onest \!\cdot I_2$ is the column-wise sum of the matrix $I_2$, which is a row vector. However, the operator in the substitution subgraph is \texttt{row\_sum} because the input to the block program is $I_2^{\!\textsf{T}}$ (the transpose matrix). Multiplying the column vector $c$ (on the left) with the columnwise-sum of $I_2$ (on the right) is another outer product. 

Like \cref{rule:scale_dot}, this rule moves the matrix multiplication operation to the first position. Another similarity with \cref{rule:scale_dot} is that all the map operators in the substitution are over the same dimension $N$, which could create opportunities for fusion.

The matched pattern can be represented with code: 
\begin{lstlisting}[style=simplestyle]
forall k in range(K):
	t1 = load(I1[k])
	t2 = t1+c[:,newaxis]
	store(t2,I4[k])
forall n in range(N):
	t3 = 0
	for k in range(K):
		t4 = load(I4[k])
		t5 = load(I2[k,n])
		t3 += t4@t5	
	store(t3,I3[n])

\end{lstlisting}
As before, \texttt{c} is a column vector in local memory. The code for the substitution is:
\begin{lstlisting}[style=simplestyle]
forall n in range(N):
	t1 = 0
	for k in range(K):
		t2 = load(I1[k])
		t3 = load(I2[k,n])
		t1 += t2@t3	
	store(t1,I4[n])
forall n in range(N):
	t4 = 0
	for k in range(K):
		t5 = load(I2[k,n])
		t4 += sum(t5, axis=0) 
	store(t4,I5[n])
forall n in range(N):
	t6 = load(I4[n])
	t7 = load(I5[n])
	t8 = outer(c,t7)+t6
	store(t8,I3[n])

\end{lstlisting}

\begin{minipage}{\columnwidth}
\addrule{Extend Map to the Entire Graph}{rule:extend}
\begin{nscenter}
\begin{tabular}{l c c c}
&Pattern:&&Substitution:\\[6pt]
(a)&\includegraphics[valign=t]{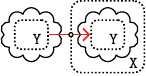}&&\includegraphics[valign=t]{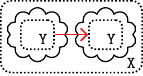}\\[30pt]
(b)&\includegraphics[valign=t]{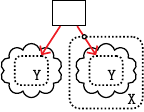}&&\includegraphics[valign=t]{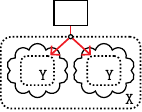}\\[50pt]
(c)&\includegraphics[valign=t]{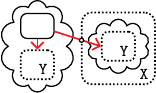}&&\includegraphics[valign=t]{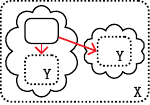}\\
\end{tabular}
\end{nscenter}
\end{minipage} \\

Imagine that the graph contains two map operators: one over dimension $Y$ and the other over dimension $X$. These operators cannot be fused because their dimensions don't match. However, consider the case where the $X$-map contains an inner graph that contains a $Y$-map. This rule extends the $X$-map to the entire graph, pulling the outer $Y$-map into its inner graph. Once both $Y$-maps are in the inner graph, they can be fused with a subsequent application of a fusion rule. 

This rule has a few different variants: variant (a) mirrors \cref{rule:consecutive} while variants (b)-(c) mirror \cref{rule:sibling}. In the first variant, the outer $Y$-map produces an intermediate result that is consumed by the inner $Y$-map. The substitution extends the $X$-map to the entire graph, and creates an opportunity to fuse the two $Y$-maps with \cref{rule:consecutive}. In the second variant, the two $Y$-maps consume the same input node, which is one of the original inputs of the block program. In the third variant, the two $Y$-maps share a common parent, which is not an input node. The second and third variants create opportunities to fuse the two $Y$-maps with \cref{rule:sibling}. 

This rule is more aggressive than the previous rules because it replicates work, which could be substantial. While this redundancy may not always increase the end-to-end execution time of the program, we still need to use this rule carefully. In the next section, where we present the actual algorithm that applies these rules, we will always create a snapshot of the block program before applying this rule, so that we can undo the work replication if it hurts performance.

\begin{minipage}{\columnwidth}
\addrule{Peel Off First Iteration}{rule:peel}  
\begin{nscenter}
\begin{tabular}{l c c}
&Pattern:&Substitution:\\[6pt]
(a)&\includegraphics[valign=t]{build/rule16}&\includegraphics[valign=t]{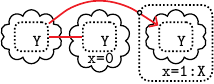}\\[30pt]
(b)&\includegraphics[valign=t]{build/rule9}&\includegraphics[valign=t]{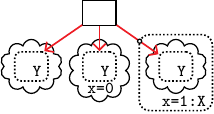}\\[50pt]
(c)&\includegraphics[valign=t]{build/rule3}&\includegraphics[valign=t]{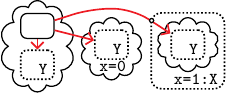}\\
\end{tabular}
\end{nscenter}
\end{minipage} \\

This rule is an alternative to \cref{rule:extend}, when redundant work is discouraged. Instead of pulling the entire graph into a map, the rule peels off the first iteration ($x=0$) and performs it separately from the rest of the map. The operator annotated with \texttt{x=0} is a copy of the map's inner graph, with the iteration variable set to $0$. The operator annotated with \texttt{x=1:X} represents the remaining $X-1$ iterations of the original map.

\begin{minipage}{\columnwidth}
\addrule{Duplicate Mapped Scale}{rule:duplicate}
\begin{nscenter}
\begin{tabular}{c c}
Pattern:&Substitution:\\[6pt]
\includegraphics[valign=t]{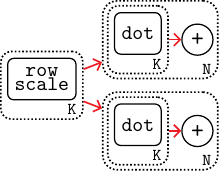}&\includegraphics[valign=t]{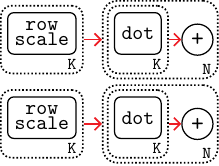}\\
\end{tabular}
\end{nscenter}
\end{minipage}

This rule duplicates a mapped scale operator that comes before two matrix multiplication operations. Applying this substitution rule enables a subsequent application of \cref{rule:scale_dot}. One could imagine a similar rule with a shift instead of a scale, to enable \cref{rule:shift_dot}.

\begin{minipage}{\columnwidth}
\addrule{Fuse Consecutive Elementwise}{rule:elementwise}
\begin{nscenter}
\begin{tabular}{c c}
Pattern:&Substitution:\\[6pt]
\includegraphics[valign=t]{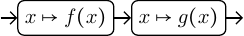}&\includegraphics[valign=t]{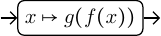}\\
\end{tabular}
\end{nscenter}
\end{minipage}

This rule fuses two consecutive elementwise operators into a single elementwise operator. It doesn't remove a large materialized intermediate result, but it does reduce the number of individual kernel invocations. Fusing elementwise operators can also have lower level benefits that exceed the scope of this paper.

\section{Algorithm} \label{sec:algo}
Recall that our rule-based fusion algorithm works hand-in-hand with a candidate selection algorithm. The input to the entire procedure is an array program, which is a directed acyclic graph whose nodes include standard array operators and custom operators. The first step is to convert this program into our block program representation, as described in \cref{sec:rep}. 

Next, the selection algorithm selects candidates and sends them to the fusion algorithm. We assume that these candidates do not contain any custom operators. If the entire block program is entirely made up of standard operators then the entire program can be one of the candidates. The fusion algorithm operates on each candidate independently and attempts to fuse it as aggressively as possible, without concern for excessive fusion. The fusion algorithm can return multiple fused implementations of each candidate and the selection algorithm will consider including each one in the final implementation.

It is important to first convert the entire array program to a block program and only then choose fusion candidates, because each array operator could be represented by multiple block operators and the best fusion candidates may not include all of them.

The fusion algorithm receives a candidate and treats it like a standalone block program, whose inputs and outputs are in global memory. The algorithm incrementally applies substitution rules from \cref{sec:sub} in a particular order and saves snapshots of some partially fused programs encountered along the way. The algorithm terminates when no more rule can be applied and returns the snapshots to the selection algorithm. 

\subsection{Fusion Without Map Extension} \label{subsec:safefuse}
Recall that the block program is a hierarchical graph and that each map operator contains an inner graph. First, we define the procedure that operates on only one of these graphs, say, the top-level graph. This procedure applies all the rules except for \cref{rule:extend}, which is applied separately. The rules are applied in a specific priority order: 
$$
\ref{rule:duplicate} \rightarrow \ref{rule:scale_dot} \rightarrow \ref{rule:shift_dot} \rightarrow \ref{rule:elementwise} \rightarrow \ref{rule:mapreduce} \rightarrow \ref{rule:consecutive} \rightarrow \ref{rule:sibling}
$$
Note that the fusion rules (1, 2, 3) appear after the companion rules. Here is the pseudocode for this procedure:
\begin{lstlisting}[style=simplestyle, tabsize=3]
def fuse_no_extend(G):
	while(True):
		M=rule8_match(G) or
			rule4_match(G) or
			rule5_match(G) or
			rule9_match(G) or
			rule3_match(G) or
			rule1_match(G) or
			rule2_match(G) 
		if not M:
			return G	
		G=apply(M)

\end{lstlisting}
We apply \texttt{fuse\_no\_extend} to each graph in breadth-first order, starting with the top-level graph, continuing to its inner graphs, and so on. Here is the pseudocode that does this:
\begin{lstlisting}[style=simplestyle, tabsize=3]
def get_inner_graphs(G):
	return [node.inner for node in G.nodes if node.type=="map"]

def bfs_fuse_no_extend(G):
	G=fuse_no_extend(G)
	queue=get_inner_graphs(G)
	while queue:
		C=queue.pop()
		C=fuse_no_extend(C)
		C.parent.inner=C
		queue.append(get_inner_graphs(C))
	return G
\end{lstlisting}
%

\subsection{Map Extension} \label{subsec:unsafefuse}
Next, we define a procedure that finds the first opportunity to apply \cref{rule:extend}. Recall that this rule extends the scope of a map node and replicates work. As before, we traverse the hierarchical graph in breadth-first order, in search of an opportunity to apply the rule. 
\begin{lstlisting}[style=simplestyle, tabsize=3]
def bfs_extend(G):
	if(M:=rule6_match(G)):
		G=apply(M)
		return G
	queue=get_inner_graphs(G)
	while queue:
		I=queue.pop()
		if(M:=rule6_match(I)):
			I=apply(M)
			I.parent.inner=I
			return G
		queue.append(get_inner_graphs(I))
	return None
\end{lstlisting}
%

\subsection{Putting it All Together} \label{subsec:together}
We run \texttt{bfs\_fuse\_no\_extend} and take a snapshot of the result. Then we run \texttt{bfs\_extend} and check if a map node was extended -- if so, we again apply \texttt{bfs\_fuse\_no\_extend} and take a snapshot. We repeat this process until \texttt{bfs\_extend} does not find a map node to extend, which implies that no more matches exist in the entire block program. 
\begin{lstlisting}[style=simplestyle, tabsize=3]
def fuse(G):
	G=bfs_fuse_no_extend(G)
	snapshots=[G]
	while(G:=bfs_extend(G)):
		G=bfs_fuse_no_extend(G)
		snapshots.append(G)
	return snapshots


	
\end{lstlisting}
%

\section{Examples} \label{sec:examples}
We demonstrate the power of our algorithm on a few interesting examples. The first example is more of a sanity check: we show that our algorithm automatically finds the celebrated Flash Attention implementation of Attention (albeit, without numerical stabilization, which we can add after the fact). 

The second and third examples are novel, and show that our algorithm extends beyond rediscovering Flash Attention. The second example involves fusing LayerNorm with matrix multiplication, which one could call \emph{Flash-LayerNorm+Matmul}. The third example is RMSNorm fused with an entire FFN-SwiGLU subgraph, which results in a mega-kernel that one could call \emph{Flash-RMSNorm+FFN-SwiGLU}. 

\onecolumn

\subsection*{Example 1: Rediscovering (Unsafe) Flash Attention}
Flash Attention uses a combination of two separate techniques: an operator fusion technique and a software-floating-point technique that adds numerical safety (sometimes referred to as \emph{online softmax}). In this section, we focus only on operator fusion and in the appendix we describe how to add numerical safety. 

\paragraph{Flash Attention, Original Array Program:}~\\ \\
\centerline{\includegraphics{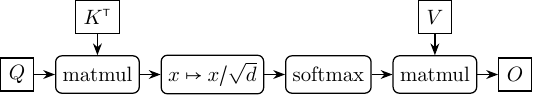}} \\ \\
This array program represents a naive implementation of the Attention operation. The input matrices $Q$ and $K^{\textsf{\tiny T}}$ are multiplied, the result is divided by the constant $\sqrt{d}$ (where $d$ is the number of columns in $Q$ and $K$), the result is passed through a softmax function, and the output is multiplied by a third matrix $V$. \\

\paragraph{Flash Attention, Initial Block Program:} We convert each array operator into its predefined subgraph of block operators. Each of the matrix multiplication operators becomes a single block operator while the softmax becomes four block operators in the top-level graph. \\ \\
\centerline{\includegraphics{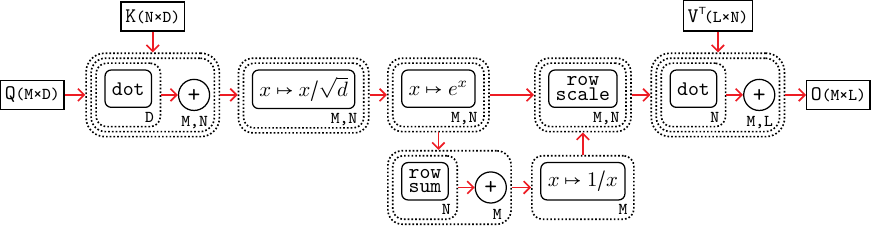}}\\ \\
Note that $K$ and $V$ are transposed in the block program, compared to the array program, because of how we defined the \texttt{dot} block operator. Also note that the original formulation of the Attention operation assumes that $Q$, $K$, and $V$ all have the same number of columns, but our block program does not assume that they are necessarily split into the same number of blocks, which is why $D$ and $L$ are separate parameters.  

\paragraph{Flash Attention, Steps 1-6: Fuse $M$-Maps} The breadth-first search starts with the top-level graph, where all the operators are $M$-dimension maps. The algorithm starts by applying \cref{rule:consecutive} or \cref{rule:sibling} six times to fuse all of them into a single $M$-map. \\ \\
\centerline{\includegraphics{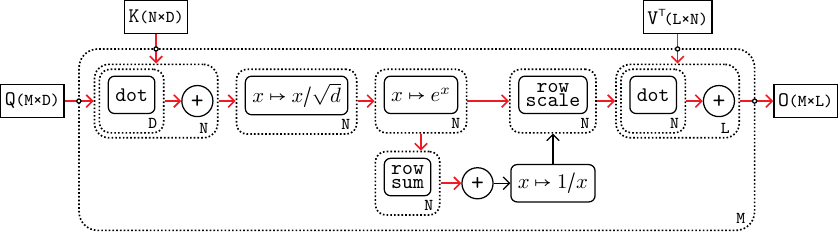}}

\begin{minipage}{\textwidth} 
This block-program represented in code:
\begin{lstlisting}[multicols=2,numbers=left,xleftmargin=17pt]
forall m in range(M):
	forall n in range(N):
		forall d in range(D):
			t1 = load(Q[m,d])
			t2 = load(KT[n,d])
			t3 = dot(t1,t2)
			store(t3,I1[m,n])
		for d in range(D):
			t4 = load(I1[m,n])
			t5 += t4
		store(t5,I2[m,n])
	forall n in range(N):
		t6 = load(I2[m,n])
		t7 = t6*(DD**-0.5)
		store(t7,I3[m,n])
	forall n in range(N):
		t8 = load(I3[m,n])
		t9 = exp(t8)
		store(t9,I4[m,n])
	forall n in range(N):
		t10 = load(I4[m,n])
		t11 = row_sum(t10)
		store(t11,I5[m,n])
	for n in range(N):
		t12 = load(I5[m,n])
		t13 += t12 
	t14 = 1/t13 
	forall n in range(N):
		t15 = load(I4[m,n])
		t16 = row_scale(t15,t14)
		store(t16,I6[m,n])
	forall l in range(L):
		forall n in range(D):
			t17 = load(I6[m,n])
			t18 = load(VT[l,n])
			t19 = dot(t17,t18)
			store(t19,I7[m,l,n])
		for n in range(D):
			t20 = load(I7[m,l,n])
			t21 += t20
		store(t21,O[m,l])
\end{lstlisting}
\end{minipage} 

The variable \texttt{DD} is a constant. For simplicity, the code implements each map as a (parallelizable) \texttt{forall} loop and each reduction as a (serial) \texttt{for} loop. The code assumes that all local variables are automatically initialized to zero, for example, \texttt{t5} on line 10 equals zero before it starts accumulating other values. 

Note that local variables in this code are scoped to the body of \texttt{forall} loops but not to the body of \texttt{for} loops. Namely, a local variable that appears inside a \texttt{forall} loop cannot be used outside of that loop. Moreover, since \texttt{forall} loops are parallelizable, each loop iteration has its own instance of the variable. For example, \texttt{t1} on line 4 exists independently for each value of $m, n$ and $d$, and can only be used in the scope of the loop (lines 4-7). On the other hand, local variables that appear inside a \texttt{for} loop continue to exist outside of that loop. For example, \texttt{t5} appears on line 10 inside a \texttt{for} loop and is used outside of that loop on line 11. There is one instance of \texttt{t5} for all $d$, but there is a separate instance of \texttt{t5} for each $m$ and $n$.

\paragraph{Flash Attention, Step 7: Swap Scale and Dot} There is nothing more to do in the top-level graph, so the breadth-first search moves to the inner graph. The algorithm now identifies an opportunity to apply \cref{rule:scale_dot}, which matches a mapped scale operation followed by a mapped dot operation. \\ \\
\centerline{\includegraphics{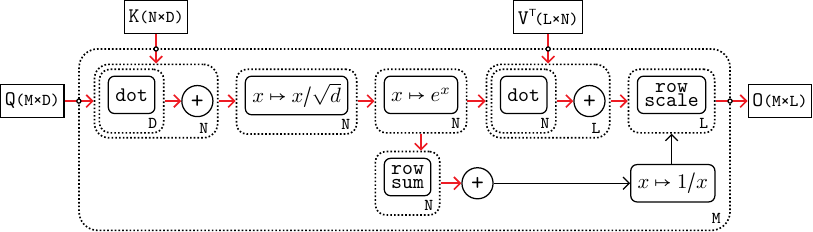}}

\begin{minipage}{\textwidth}
This block-program represented in code:
\begin{lstlisting}[multicols=2,numbers=left,xleftmargin=17pt]
forall m in range(M):
	forall n in range(N):
		forall d in range(D):
			t1 = load(Q[m,d])
			t2 = load(KT[n,d])
			t3 = dot(t1,t2)
			store(t3,I1[m,n])
		for d in range(D):
			t4 = load(I1[m,n])
			t5 += t4
		store(t5,I2[m,n])
	forall n in range(N):
		t6 = load(I2[m,n])
		t7 = t6*(DD**-0.5)
		store(t7,I3[m,n])
	forall n in range(N):
		t8 = load(I3[m,n])
		t9 = exp(t8)
		store(t9,I4[m,n])
	forall l in range(L):
		forall n in range(D):
			t10 = load(I4[m,n])
			t11 = load(VT[l,n])
			t12 = dot(t10,t11)
			store(t12,I5[m,l,n])
		for n in range(D):
			t13 = load(I5[m,l,n])
			t14 += t13
		store(t14,I6[m,l])
	forall n in range(N):
		t15 = load(I4[m,n])
		t16 = row_sum(t15)
		store(t16,I7[m,n])
	for n in range(N):
		t17 = load(I7[m,n])
		t18 += t17 
	t19 = 1/t18 
	forall l in range(L):
		t20 = load(I6[m,l])
		t21 = row_scale(t20,t19)
		store(t21,O[m,l])
\end{lstlisting}
\end{minipage}

\paragraph{Flash Attention, Step 8: Fuse Map and Reduction} Still on the same graph graph, the algorithm identifies an opportunity to apply \cref{rule:mapreduce}, which fuses a map operator with a reduction. \\ \\
\centerline{\includegraphics{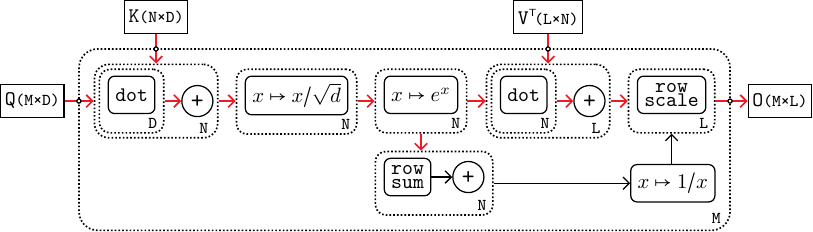}}

\begin{minipage}{\textwidth}
This block-program represented in code:
\begin{lstlisting}[multicols=2,numbers=left,xleftmargin=17pt]
forall m in range(M):
	forall n in range(N):
		forall d in range(D):
			t1 = load(Q[m,d])
			t2 = load(KT[n,d])
			t3 = dot(t1,t2)
			store(t3,I1[m,n])
		for d in range(D):
			t4 = load(I1[m,n])
			t5 += t4
		store(t5,I2[m,n])
	forall n in range(N):
		t6 = load(I2[m,n])
		t7 = t6*(DD**-0.5)
		store(t7,I3[m,n])
	forall n in range(N):
		t8 = load(I3[m,n])
		t9 = exp(t8)
		store(t9,I4[m,n])
	forall l in range(L):
		forall n in range(D):
			t10 = load(I4[m,n])
			t11 = load(VT[l,n])
			t12 = dot(t10,t11)
			store(t12,I5[m,l,n])
		for n in range(D):
			t13 = load(I5[m,l,n])
			t14 += t13
		store(t14,I6[m,l])
	for n in range(N):
		t15 = load(I4[m,n])
		t16 += row_sum(t15)
	t17 = 1/t16 
	forall l in range(L):
		t18 = load(I6[m,l])
		t19 = row_scale(t18,t17)
		store(t19,O[m,l])
\end{lstlisting}
\end{minipage}\\ \\

\paragraph{Flash Attention, Step 9-12: Fuse $N$-maps and $L$-maps} Still on the same graph, the algorithm applies \cref{rule:consecutive} three times to fuse all the $N$-maps, and one more time to fuse the two $L$-maps.\\ \\
\centerline{\includegraphics{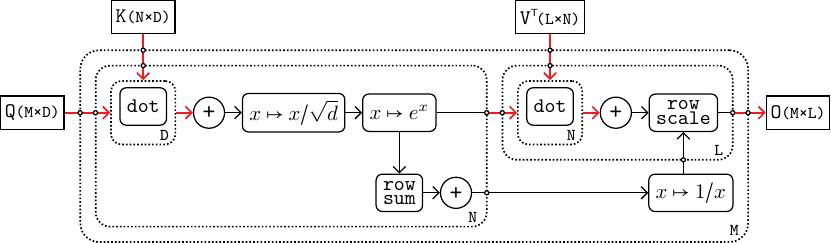}}

\begin{minipage}{\textwidth}
This block-program represented in code:
\begin{lstlisting}[multicols=2,numbers=left,xleftmargin=17pt]
forall m in range(M):
	for n in range(N):
		forall d in range(D):
			t1 = load(Q[m,d])
			t2 = load(KT[n,d])
			t3 = dot(t1,t2)
			store(t3,I1[m,n])
		for d in range(D):
			t4 = load(I1[m,n])
			t5 += t4
		t6 = t5*(DD**-0.5)
		t7 = exp(t6)
		store(t7,I4[m,n])
		t8 += row_sum(t7)
	t9 = 1/t8 
	forall l in range(L):
		forall n in range(D):
			t10 = load(I4[m,n])
			t11 = load(VT[l,n])
			t12 = dot(t10,t11)
			store(t12,I5[m,l,n])
		for n in range(D):
			t13 = load(I5[m,l,n])
			t14 += t13
		t15 = row_scale(t14,t9)
		store(t15,O[m,l])
\end{lstlisting}
\end{minipage}

\paragraph{Flash Attention, Step 13: Fuse Elementwise} There is nothing more to do in this graph, so the breadth-first search proceeds to the inner graph of the $N$-map. There, the algorithm applies \cref{rule:elementwise}, which fuses two consecutive elementwise operations. \\ \\
\centerline{\includegraphics{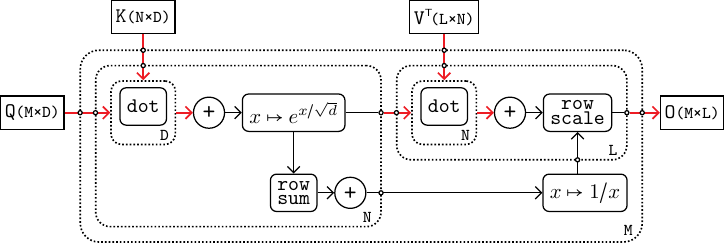}}

\begin{minipage}{\textwidth}
This block-program represented in code:
\begin{lstlisting}[multicols=2,numbers=left,xleftmargin=17pt]
forall m in range(M):
	for n in range(N):
		forall d in range(D):
			t1 = load(Q[m,d])
			t2 = load(KT[n,d])
			t3 = dot(t1,t2)
			store(t3,I1[m,n])
		for d in range(D):
			t4 = load(I1[m,n])
			t5 += t4
		t6 = exp(t5*(DD**-0.5))
		store(t6,I4[m,n])
		t7 += row_sum(t6)
	t8 = 1/t7 
	forall l in range(L):
		forall n in range(D):
			t9 = load(I4[m,n])
			t10 = load(VT[l,n])
			t11 = dot(t9,t10)
			store(t11,I5[m,l,n])
		for n in range(D):
			t12 = load(I5[m,l,n])
			t13 += t12
		t14 = row_scale(t13,t8)
		store(t14,O[m,l])
\end{lstlisting}
\end{minipage}

\paragraph{Flash Attention, Step 14-15: Fuse Maps with Reductions} Still in the same graph, the algorithm applies \cref{rule:mapreduce}, which fuses the $D$-map with a consecutive reduction. After that, the breadth-first search moves to the inner graph in the $L$-map, where the algorithm applies \cref{rule:mapreduce} to fuse the $N$-map with the consecutive reduction. \\ \\
\centerline{\includegraphics{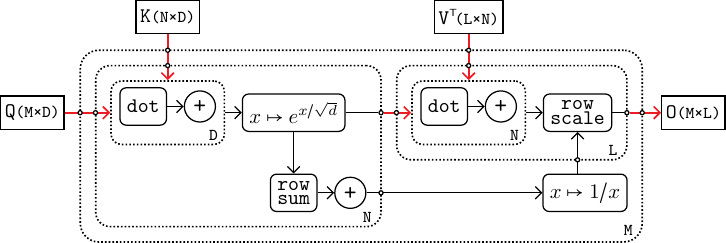}}

\begin{minipage}{\textwidth}
This block-program represented in code:
\begin{lstlisting}[multicols=2,numbers=left,xleftmargin=17pt]
forall m in range(M):
	for n in range(N):
		for d in range(D):
			t1 = load(Q[m,d])
			t2 = load(KT[n,d])
			t3 += dot(t1,t2)
		t4 = exp(t3*(DD**-0.5))
		store(t4,I4[m,n])
		t5 += row_sum(t4)
	t6 = 1/t5 
	forall l in range(L):
		for n in range(D):
			t7 = load(I4[m,n])
			t8 = load(VT[l,n])
			t9 += dot(t7,t8)
		t10 = row_scale(t9,t6)
		store(t10,O[m,l])
\end{lstlisting}
\end{minipage}

\paragraph{Flash Attention, Step 16: Extend the $L$-map} At this point, the \texttt{bfs\_fuse\_no\_extend} procedure has run its course. The top-level procedure \texttt{fuse} captures a snapshot of the block-program, and then calls \texttt{bfs\_extend}, which uses \cref{rule:extend} to extend the $L$-map to the entire inner graph. \\ \\
\centerline{\includegraphics{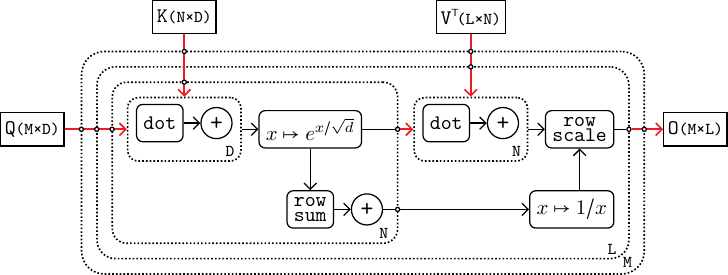}}

\begin{minipage}{\textwidth}
This block-program represented in code:
\begin{lstlisting}[multicols=2,numbers=left,xleftmargin=17pt]
forall m in range(M):
	forall l in range(L):
		for n in range(N):
			for d in range(D):
				t1 = load(Q[m,d])
				t2 = load(KT[n,d])
				t3 += dot(t1,t2)
			t4 = exp(t3*(DD**-0.5))
			store(t4,I4[m,n])
			t5 += row_sum(t4)
		t6 = 1/t5 
		for n in range(D):
			t7 = load(I4[m,n])
			t8 = load(VT[l,n])
			t9 += dot(t7,t8)
		t10 = row_scale(t9,t6)
		store(t10,O[m,l])
\end{lstlisting}
\end{minipage}

\paragraph{Flash Attention, Step 17: Fuse Consecutive $N$-maps} Finally, the algorithm fuses the two consecutive $N$-maps using \cref{rule:consecutive}.\\ \\
\centerline{\includegraphics{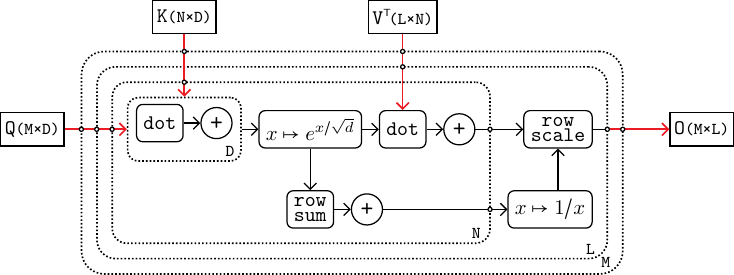}}

\begin{minipage}{\textwidth}
This block-program represented in code:
\begin{lstlisting}[multicols=2,numbers=left,xleftmargin=17pt]
forall m in range(M):
	forall l in range(L):
		for n in range(N):
			for d in range(D):
				t1 = load(Q[m,d])
				t2 = load(KT[n,d])
				t3 += dot(t1,t2)
			t4 = exp(t3*(DD**-0.5))
			t5 += row_sum(t4)
			t6 = load(VT[l,n])
			t7 += dot(t4,t6)
		t8 = 1/t5 
		t9 = row_scale(t7,t8)
		store(t9,O[m,l])
\end{lstlisting}
\end{minipage}\\ \\

At this point, the fusion algorithm has exhausted all opportunities to match rules against any of the graphs.  

\paragraph{Flash Attention, Epilogue:} The only remaining buffered edges are those that are incident with input or output nodes, which means that the algorithm has successfully fused the entire program. Recall that we mentioned that the parameters $M,D,N,L$ can be autotuned -- in particular, the autotuner will consider setting $D=L=1$, which are the values that reproduce the original Flash Attention kernel. 

Additional fusion steps may be possible, but they depend on specific characteristics of the target hardware and exceed the scope of this paper. For example, on a CPU target, the block operators are themselves implemented as loops, which can be further fused (e.g., one could fuse lines 8,10,11 and lines 12,13). 


\subsection*{Example 2: Fusing LayerNorm and Matmul}
LayerNorm is an operation that normalizes each row of a matrix by subtracting its mean and then dividing by its standard deviation. Consider an array program that applies a LayerNorm operation to a matrix $X$ and then multiplies the result by another matrix $Y$.

\paragraph{LayerNorm+Matmul, Original Array Program:}~\\
\centerline{\includegraphics{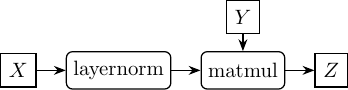}}

\paragraph{LayerNorm+Matmul, Initial Block Program:}  We convert each array operator into its predefined subgraph of block operators. The LayerNorm operator becomes multiple block operators, while matrix multiplication becomes a single block operator in the top-level graph. \\ \\
\centerline{\includegraphics{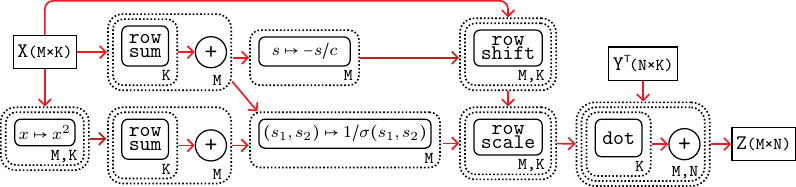}}\\ \\
The program starts by calculating the sum and the sum-of-squares of each row in the matrix. The sum is used to calculate the mean of each row. The standard deviation of each row is calculated using the formula 
\begin{equation}\label{eqn:std}
\sigma(s_1,s_2) = \sqrt{\frac{s_2}{k} - \left(\frac{s_1}{k}\right)^2}~~,
\end{equation}
where $s_1$ is the sum, $s_2$ is the sum-of-squares, and $k$ is the number of elements being summed. The program then subtracts the row-mean from each element in the row and divides the result by the row-standard-deviation. Finally, the program multiplies the result by the matrix $Y$. Note that the program takes $Y^\top$ because of how we defined the \texttt{dot} block operator.

\paragraph{LayerNorm+Matmul, Steps 1-7: Fuse $M$-Maps} The breadth-first search starts with the top-level graph, where all the operators are $M$-dimension maps. The algorithm starts by applying \cref{rule:consecutive} or \cref{rule:sibling} seven times to fuse all of them into a single $M$-map. \\ \\
\centerline{\includegraphics{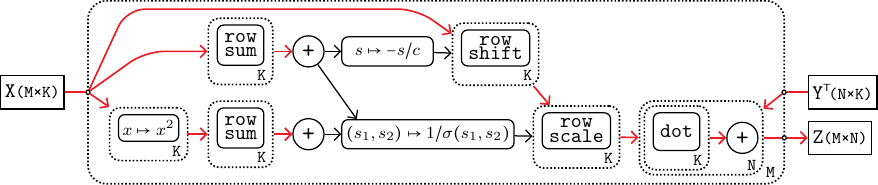}}

\begin{minipage}{\textwidth}
This program can also be represented using code:
\begin{lstlisting}[multicols=2,numbers=left,xleftmargin=17pt]
forall m in range(M):
	forall k in range(K):
		t1 = load(X[m,k])
		t2 = row_sum(t1)
		store(t2,I1[m,k])
	for k in range(K):
		t3 = load(I1[m,k])
		t4 += t3
	t5 = -t4/KK
	forall k in range(K):
		t6 = load(X[m,k])
		t7 = row_shift(t6,t5)
		store(t7,I2[m,k])
	forall k in range(K):
		t8 = load(X[m,k])
		t9 = t8**2
		store(t9,I3[m,k])
	forall k in range(K):
		t10 = load(I3[m,k])
		t11 = row_sum(t10)
		store(t11,I4[m,k])
	for k in range(K):
		t12 = load(I4[m,k])
		t13 += t12
	t14 = (t13/KK-t5**2)**(-0.5)
	forall k in range(K):
		t15 = load(I2[m,k])
		t16 = row_scale(t15,t14)
		store(t16,I5[m,k])
	forall n in range(N):
		forall k in range(K):
			t17 = load(I5[m,k])
			t18 = load(YT[n,k])
			t19 = dot(t17,t18)
			store(t19, I6[m,n,k])
		for k in range(K):
			t20 = load(I6[m,n,k])
			t21 += t20
		store(t21, Z[m,n])
\end{lstlisting} 
\end{minipage}

The variable \texttt{KK} in the code above is the same as $k$ in \cref{eqn:std}, which equals the number of columns in $X$. The code above follows the same conventions (reductions as for loops, scope, etc.) described in the previous example. 

\paragraph{LayerNorm+Matmul, Step 8: Swap Scale and Dot} There is nothing more to do in the top-level graph, so the breadth-first search moves to the inner graph. There, the algorithm applies \cref{rule:scale_dot}, which matches a mapped scale operation followed by a mapped dot operation. \\ \\
\centerline{\includegraphics{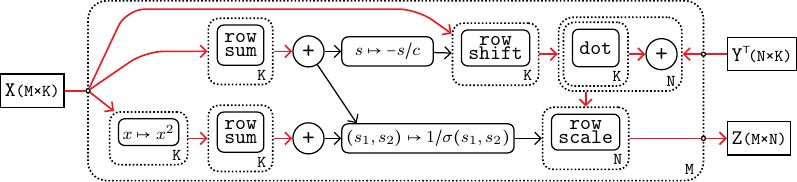}}

\begin{minipage}{\textwidth}
This block-program represented in code:
\begin{lstlisting}[multicols=2,numbers=left,xleftmargin=17pt]
forall m in range(M):
	forall k in range(K):
		t1 = load(X[m,k])
		t2 = row_sum(t1)
		store(t2,I1[m,k])
	for k in range(K):
		t3 = load(I1[m,k])
		t4 += t3
	t5 = -t4/KK
	forall k in range(K):
		t6 = load(X[m,k])
		t7 = row_shift(t6,t5)
		store(t7,I2[m,k])
	forall n in range(N):
		forall k in range(K):
			t8 = load(I2[m,k])
			t9 = load(YT[n,k])
			t10 = dot(t8,t9)
			store(t10, I3[m,n,k])
		for k in range(K):
			t11 = load(I3[m,n,k])
			t12 += t11
		store(t12, I4[m,n])
	forall k in range(K):
		t12 = load(X[m,k])
		t13 = t12**2
		store(t13,I5[m,k])
	forall k in range(K):
		t14 = load(I5[m,k])
		t15 = row_sum(t14)
		store(t15,I6[m,k])
	for k in range(K):
		t16 = load(I6[m,k])
		t17 += t16
	t18 = (t17/KK-t5**2)**(-0.5)
	forall n in range(N):
		t19 = load(I4[m,n])
		t20 = row_scale(t19,t18)
		store(t20,Z[m,n])
\end{lstlisting}
\end{minipage}

\paragraph{LayerNorm+Matmul, Step 9: Swap Shift and Dot} Still focused on the same graph, the algorithm applies \cref{rule:shift_dot}, which matches a mapped scale operation followed by a mapped dot operation. \\ \\
\centerline{\includegraphics{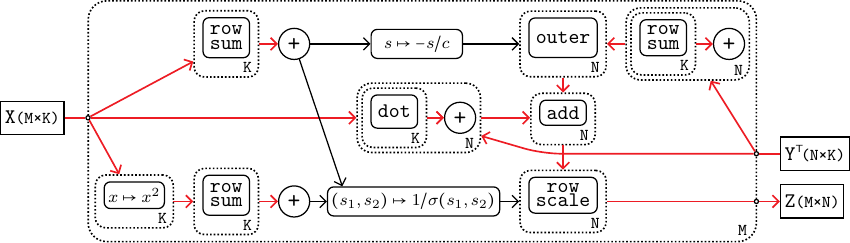}}

\begin{minipage}{\textwidth}
This block-program represented in code:
\begin{lstlisting}[multicols=2,numbers=left,xleftmargin=17pt]
forall m in range(M):
	forall k in range(K):
		t1 = load(X[m,k])
		t2 = row_sum(t1)
		store(t2,I1[m,k])
	for k in range(K):
		t3 = load(I1[m,k])
		t4 += t3
	t5 = -t4/KK
	forall n in range(N):
		forall k in range(K):
			t6 = load(YT[n,k])
			t7 = row_sum(t6)
			store(t7,I2[m,n,k])
		forall k in range(K):
			t8 = load(I2[m,n,k])
			t9 += t8
		store(t9,I3[m,n])
	forall n in range(N):
		t10 = load(I3[m,n])
		t11 = outer(t5,t10)
		store(t11,I4[m,n])
	forall n in range(N):
		forall k in range(K):
			t12 = load(X[m,k])
			t13 = load(YT[n,k])
			t14 = dot(t12,t13)
			store(t14, I5[m,n,k])
		for k in range(K):
			t15 = load(I5[m,n,k])
			t16 += t15
		store(t16,I6[m,n])
	forall n in range(N):
		t17 = load(I4[m,n])
		t18 = load(I6[m,n])
		t19 = add(t17,t18)
		store(t19,I7[m,n])
	forall k in range(K):
		t20 = load(X[m,k])
		t21 = t20**2
		store(t21,I8[m,k])
	forall k in range(K):
		t22 = load(I8[m,k])
		t23 = row_sum(t22)
		store(t23,I9[m,k])
	for k in range(K):
		t24 = load(I9[m,k])
		t25 += t24
	t26 = (t25/KK-t5**2)**(-0.5)
	forall n in range(N):
		t27 = load(I7[m,n])
		t28 = row_scale(t27,t26)
		store(t28,Z[m,n])
\end{lstlisting}
\end{minipage}

At this point, the code doesn't appear to be any better than the code we started out with. However, while it may not be immediately apparent, the algorithm's last two steps have exposed opportunities that the next steps will take advantage of.  

\paragraph{LayerNorm+Matmul, Steps 10-11: Fusing Maps with Reductions} Still focused on the same graph, the algorithm identifies two separate opportunities to fuse map operators with adjacent reduction operators using \cref{rule:mapreduce}. \\ \\ 
\centerline{\includegraphics{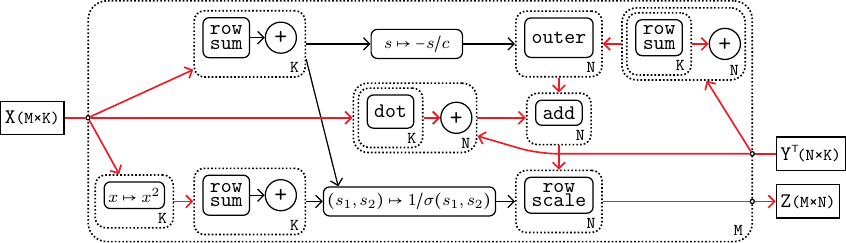}}

\begin{minipage}{\textwidth}
This block-program represented in code:
\begin{lstlisting}[multicols=2,numbers=left,xleftmargin=17pt]
forall m in range(M):
	for k in range(K):
		t1 = load(X[m,k])
		t2 += row_sum(t1)
	t3 = -t2/KK
	forall n in range(N):
		forall k in range(K):
			t4 = load(YT[n,k])
			t5 = row_sum(t4)
			store(t5,I1[m,n,k])
		forall k in range(K):
			t6 = load(I1[m,n,k])
			t7 += t6
		store(t7,I2[m,n])
	forall n in range(N):
		t8 = load(I2[m,n])
		t9 = outer(t3,t8)
		store(t9,I3[m,n])
	forall n in range(N):
		forall k in range(K):
			t10 = load(X[m,k])
			t11 = load(YT[n,k])
			t12 = dot(t10,t11)
			store(t12, I4[m,n,k])
		for k in range(K):
			t13 = load(I4[m,n,k])
			t14 += t13
		store(t14,I5[m,n])
	forall n in range(N):
		t15 = load(I3[m,n])
		t16 = load(I5[m,n])
		t17 = add(t15,t16)
		store(t17,I6[m,n])
	forall k in range(K):
		t18 = load(X[m,k])
		t19 = t18**2
		store(t19,I7[m,k])
	for k in range(K):
		t20 = load(I7[m,k])
		t21 += row_sum(t20)
	t22 = (t21/KK-t3**2)**(-0.5)
	forall n in range(N):
		t23 = load(I6[m,n])
		t24 = row_scale(t23,t22)
		store(t24,Z[m,n])
\end{lstlisting}
\end{minipage}

\paragraph{LayerNorm+Matmul, Steps 12-17: Fuse $N$ and $L$ Maps} Still focused on the same graph, the algorithm applies \cref{rule:consecutive} four times to fuse all the $N$-maps, and Rules 
\ref{rule:consecutive} and \ref{rule:sibling} to fuse the $K$-maps.\\ \\
\centerline{\includegraphics{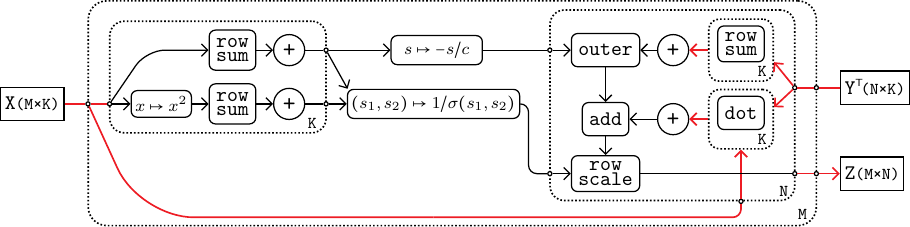}}

\begin{minipage}{\textwidth}
This block-program represented in code:
\begin{lstlisting}[multicols=2,numbers=left,xleftmargin=17pt]
forall m in range(M):
	for k in range(K):
		t1 = load(X[m,k])
		t2 += row_sum(t1)
		t3 += row_sum(t1**2)
	t4 = -t2/KK
	t5 = (t3/KK-t4**2)**(-0.5)
	forall n in range(N):
		forall k in range(K):
			t6 = load(YT[n,k])
			t7 = row_sum(t6)
			store(t7,I1[m,n,k])
		for k in range(K):
			t8 = load(I1[m,n,k])
			t9 += t8
		t10 = outer(t4,t9)
		forall k in range(K):
			t11 = load(X[m,k])
			t12 = load(YT[n,k])
			t13 = dot(t11,t12)
			store(t13, I2[m,n,k])
		for k in range(K):
			t14 = load(I2[m,n,k])
			t15 += t14
		t16 = add(t10,t15)
		t17 = row_scale(t16,t5)
		store(t17,Z[m,n])
\end{lstlisting} 
\end{minipage}

\paragraph{LayerNorm+Matmul, Step 18-19: Fuse Maps with Reductions} There is nothing left to do in the inner graph, so the breadth-first search proceeds to the next level of graphs, and specifically, to the inner graph of the $N$-map. There, the algorithm identifies two separate opportunities to fuse maps with reductions using\cref{rule:mapreduce}.\\ \\
\centerline{\includegraphics{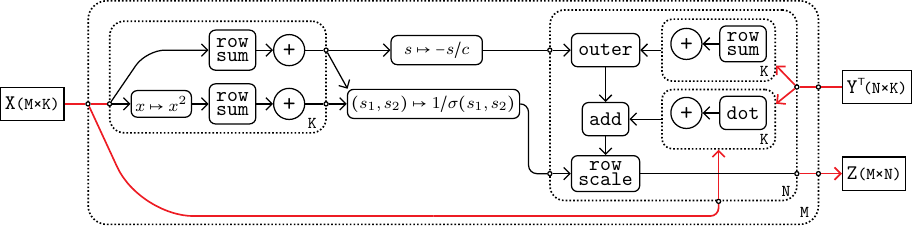}}

\begin{minipage}{\textwidth}
This block-program represented in code:
\begin{lstlisting}[multicols=2,numbers=left,xleftmargin=17pt]
forall m in range(M):
	for k in range(K):
		t1 = load(X[m,k])
		t2 += row_sum(t1)
		t3 += row_sum(t1**2)
	t4 = -t2/KK
	t5 = (t3/KK-t4**2)**(-0.5)
	forall n in range(N):
		for k in range(K):
			t6 = load(YT[n,k])
			t7 += row_sum(t6)
		t8 = outer(t4,t7)
		for k in range(K):
			t9 = load(X[m,k])
			t10 = load(YT[n,k])
			t11 += dot(t9,t10)
		t12 = add(t8,t11)
		t13 = row_scale(t12,t5)
		store(t13,Z[m,n])
\end{lstlisting}
\end{minipage} 

\paragraph{LayerNorm+Matmul, Step 20: Fuse Sibling $K$-Maps} Still focused on the same graph, the algorithm fuses the two $K$-maps using \cref{rule:sibling}\\ \\
\centerline{\includegraphics{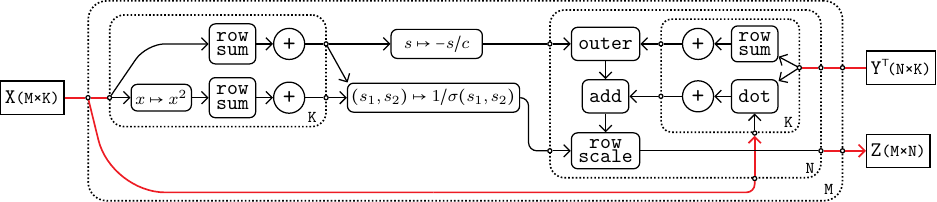}}

\begin{minipage}{\textwidth}
This block-program represented in code:
\begin{lstlisting}[multicols=2,numbers=left,xleftmargin=17pt]
forall m in range(M):
	for k in range(K):
		t1 = load(X[m,k])
		t2 += row_sum(t1)
		t3 += row_sum(t1**2)
	t4 = -t2/KK
	t5 = (t3/KK-t4**2)**(-0.5)
	forall n in range(N):
		for k in range(K):
			t6 = load(YT[n,k])
			t7 += row_sum(t6)
			t8 = load(X[m,k])
			t9 += dot(t8,t6)
		t10 = outer(t4,t7)
		t11 = add(t10,t9)
		t12 = row_scale(t11,t5)
		store(t12,Z[m,n])
\end{lstlisting}
\end{minipage}

\paragraph{LayerNorm+Matmul, Step 21: Extend the $N$-Map} The \texttt{bfs\_fuse\_no\_extend} procedure has now run out of matches at all levels of the block program. Therefore, the top-level function \texttt{fuse} captures a snapshot of the block-program and then calls \texttt{bfs\_extend}, which applies \cref{rule:extend} to extend the $N$-map to the entire inner graph. \\ \\
\centerline{\includegraphics{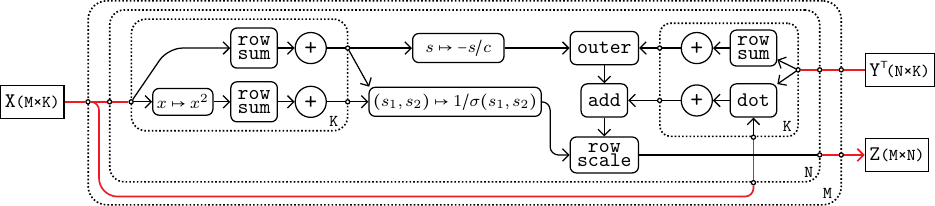}}

\begin{minipage}{\textwidth}
This block-program represented in code:
\begin{lstlisting}[multicols=2,numbers=left,xleftmargin=17pt]
forall m in range(M):
	forall n in range(N):
		for k in range(K):
			t1 = load(X[m,k])
			t2 += row_sum(t1)
			t3 += row_sum(t1**2)
		t4 = -t2/KK
		t5 = (t3/KK-t4**2)**(-0.5)
		for k in range(K):
			t6 = load(YT[n,k])
			t7 += row_sum(t6)
			t8 = load(X[m,k])
			t9 += dot(t8,t6)
		t10 = outer(t4,t7)
		t11 = add(t10,t9)
		t12 = row_scale(t11,t5)
		store(t12,Z[m,n])
\end{lstlisting}
\end{minipage}

\paragraph{LayerNorm+Matmul, Step 22: Fuse Sibling $K$-Maps} The previous step exposed an opportunity to fuse two $K$-maps using \cref{rule:sibling}.\\ \\
\centerline{\includegraphics{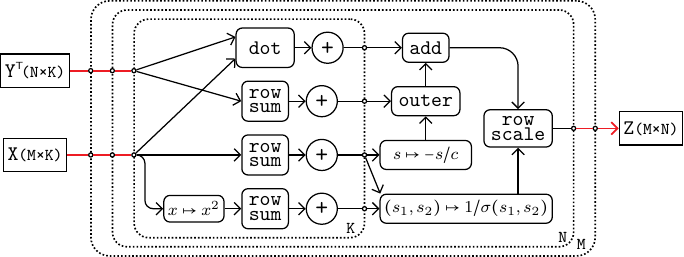}}

\begin{minipage}{\textwidth}
This block-program represented in code:
\begin{lstlisting}[multicols=2,numbers=left,xleftmargin=17pt]
forall m in range(M):
	forall n in range(N):
		for k in range(K):
			t1 = load(X[m,k])
			t2 += row_sum(t1)
			t3 += row_sum(t1**2)
			t4 = load(YT[n,k])
			t5 += row_sum(t4)
			t6 += dot(t1,t4)
		t7 = -t2/KK
		t8 = (t3/KK-t7**2)**(-0.5)
		t9 = outer(t7,t5)
		t10 = add(t9,t6)
		t11 = row_scale(t10,t8)
		store(t11,Z[m,n])
\end{lstlisting}
\end{minipage}

At this point, the fusion algorithm has exhausted all opportunities to match rules against any of the graphs.  

\paragraph{LayerNorm+Matmul, Epilogue:} The only remaining buffered edges are those that are incident with input or output nodes, which means that the algorithm has successfully fused the entire program. As in the previous example, additional fusion steps may be possible, but they depend on specific characteristics of the target hardware and exceed the scope of this paper. For example, on a CPU target, which computes individual block operators using loops, one could further fuse lines 5-6 and lines 10-14. 

\subsection*{Example 3: Fusing RMSNorm and FeedForward-SwiGLU}
In the classic Transformer architecture, the well-known Multihead Attention operator is followed by a normalization operation and an operation labeled ``feedforward''. In modern Transformers, the normalization is often an RMSNorm and the feedforward operation is the mysterious FFN-SwiGLU \citep{Shazeer2020}. The combination of RMSNorm and FFN-SwiGLU includes three matrix multiplications, a row-wise reduction, a Hadamard product (elementwise multiplication), and a few elementwise operations. This example shows how our algorithm tightly fuses all of these operations into a single mega-kernel.

Specifically, the program begins by normalizing the matrix $X$ and multiplying the result twice, by $W$ and $V$. One of the products is passed through an elementwise Swish operation. Then, we take the Hadamard product of the two matrices, and multiply the result by a third matrix $U$.

\paragraph{RMS+FNN-SwiGLU, Original Array Program:}~\\ \\
\centerline{\includegraphics{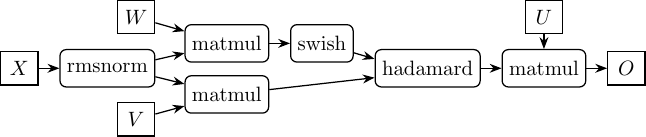}}

\paragraph{RMS+FNN-SwiGLU, Initial Block Program:} We convert each array operator into its predefined subgraph of block operators. The matrix multiplications, the Hadamard product, and the elementwise operation each become a single block operator in the top-level graph. The RMSNorm becomes a sequence of four block operators, three of them calculate the inverse root-mean-square and the fourth scales the input by it. \\ \\
\centerline{\includegraphics{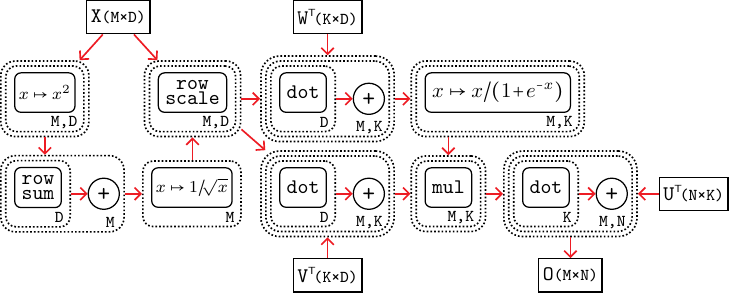}}

\paragraph{RMS+FNN-SwiGLU, Steps 1-8: Fuse $M$-Maps} The breadth-first search starts with the top-level graph, where all the operators are $M$-dimension maps. The algorithm starts by applying \cref{rule:consecutive} or \cref{rule:sibling} eight times to fuse all of them into a single $M$-map.

\centerline{\includegraphics{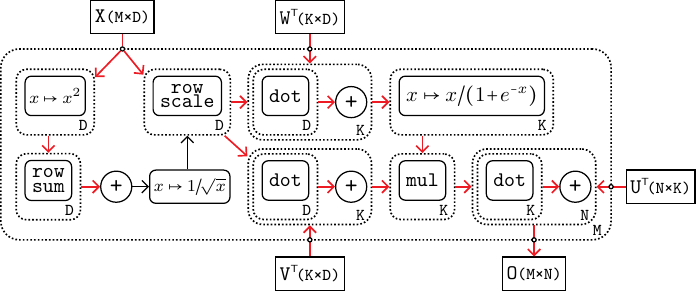}}

\begin{minipage}{\textwidth}
This block-program represented in code:
\begin{lstlisting}[multicols=2,numbers=left,xleftmargin=17pt]
forall m in range(M):
	forall d in range(D):
		t1 = load(X[m,d])
		t2 = t1**2
		store(t2,I1[m,d])
	forall d in range(D):
		t3 = load(I1[m,d])
		t4 = row_sum(t3)
		store(t4,I2[m,d])
	for d in range(D):
		t5 = load(I2[m,d])
		t6 += t5
	t7 = 1/sqrt(t6)
	forall d in range(D):
		t8 = load(X[m,d])
		t9 = row_scale(t8,t7)
		store(t9,I3[m,d])
	forall k in range(K):
		forall d in range(D):
			t10 = load(I3[m,d])
			t11 = load(WT[k,d])
			t12 = dot(t10,t11)
			store(t12,I4[m,k,d])
		for d in range(D):
			t13 = load(I4[m,k,d])
			t14 += t13
		store(t14,I5[m,k])
	forall k in range(K):
		t15 = load(I5[m,k])
		t16 = t15/(1+exp(-t15))
		store(t16,I6[m,k])
	forall k in range(K):
		forall d in range(D):
			t17 = load(I3[m,d])
			t18 = load(VT[k,d])
			t19 = dot(t17,t18)
			store(t19,I7[m,k,d])
		for d in range(D):
			t20 = load(I7[m,k,d])
			t21 += t20
		store(t21,I8[m,k])
	forall k in range(K):
		t22 = load(I6[m,k])
		t23 = load(I8[m,k])
		t24 = mul(t22,t23)
		store(t24,I9[m,k])
	forall n in range(N):
		forall k in range(K):
			t25 = load(I9[m,k])
			t26 = load(UT[n,k])
			t27 = dot(t25,t26)
			store(t27,I10[m,n,k])
		for k in range(K):
			t28 = load(I10[m,n,k])
			t29 += t28
		store(t29,O[m,n])

\end{lstlisting}
\end{minipage}

The code above follows the same conventions (reductions as for loops, scope, etc.) described in the previous examples.

\paragraph{RMS+FNN-SwiGLU, Step 9: Duplicate Row Scale} There is nothing more to do in the top-level graph, so the breadth-first search moves to the inner graph, where the algorithm duplicates the row-wise scaling using \cref{rule:duplicate}.\\ \\
\centerline{\includegraphics{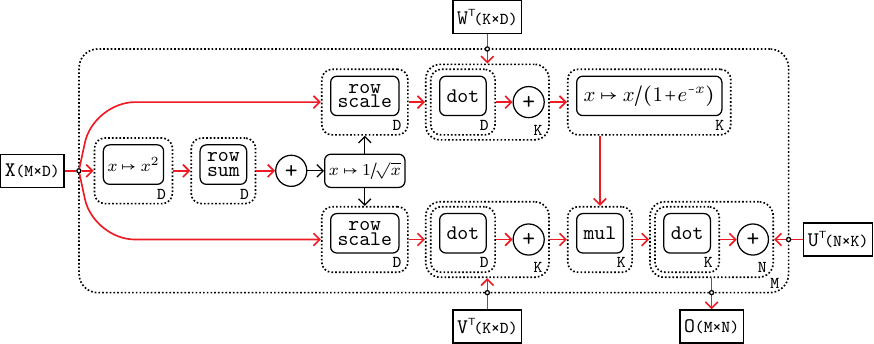}}

\begin{minipage}{\textwidth}
This block-program represented in code:
\begin{lstlisting}[multicols=2,numbers=left,xleftmargin=17pt]
forall m in range(M):
	forall d in range(D):
		t1 = load(X[m,d])
		t2 = t1**2
		store(t2,I1[m,d])
	forall d in range(D):
		t3 = load(I1[m,d])
		t4 = row_sum(t3)
		store(t4,I2[m,d])
	for d in range(D):
		t5 = load(I2[m,d])
		t6 += t5
	t7 = 1/sqrt(t6)
	forall d in range(D):
		t8 = load(X[m,d])
		t9 = row_scale(t8,t7)
		store(t9,I3[m,d])
	forall k in range(K):
		forall d in range(D):
			t10 = load(I3[m,d])
			t11 = load(WT[k,d])
			t12 = dot(t10,t11)
			store(t12,I4[m,k,d])
		for d in range(D):
			t13 = load(I4[m,k,d])
			t14 += t13
		store(t14,I5[m,k])
	forall k in range(K):
		t15 = load(I5[m,k])
		t16 = t15/(1+exp(-t15))
		store(t16,I6[m,k])
	forall d in range(D):
		t17 = load(X[m,d])
		t18 = row_scale(t17,t7)
		store(t18,I7[m,d])
	forall k in range(K):
		forall d in range(D):
			t19 = load(I7[m,d])
			t20 = load(VT[k,d])
			t21 = dot(t19,t20)
			store(t21,I8[m,k,d])
		for d in range(D):
			t22 = load(I8[m,k,d])
			t23 += t22
		store(t23,I9[m,k])
	forall k in range(K):
		t24 = load(I6[m,k])
		t25 = load(I9[m,k])
		t26 = mul(t24,t25)
		store(t26,I10[m,k])
	forall n in range(N):
		forall k in range(K):
			t27 = load(I10[m,k])
			t28 = load(UT[n,k])
			t29 = dot(t27,t28)
			store(t29,I11[m,n,k])
		for k in range(K):
			t30 = load(I11[m,n,k])
			t31 += t30
		store(t31,O[m,n])
\end{lstlisting}
\end{minipage}

\paragraph{RMS+FNN-SwiGLU, Steps 10-11: Swap Scale and Dot} The algorithm then identifies two separate opportunities to swap a mapped row-wise scaling with a matrix multiplication using \cref{rule:scale_dot}. The order in which these substitutions occur is not important.\\ \\
\centerline{\includegraphics{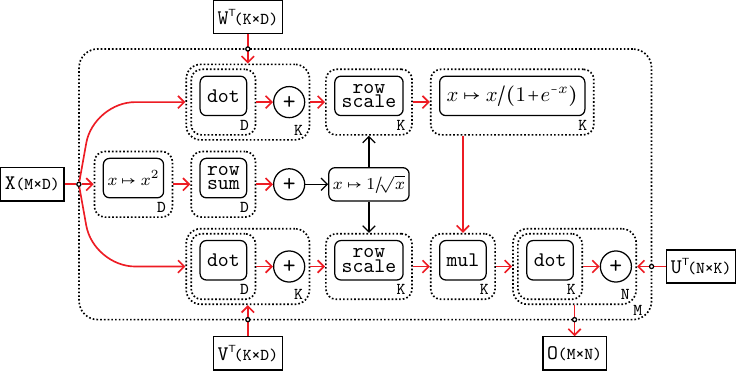}}

\begin{minipage}{\textwidth}
This block-program represented in code:
\begin{lstlisting}[multicols=2,numbers=left,xleftmargin=17pt]
forall m in range(M):
	forall d in range(D):
		t1 = load(X[m,d])
		t2 = t1**2
		store(t2,I1[m,d])
	forall d in range(D):
		t3 = load(I1[m,d])
		t4 = row_sum(t3)
		store(t4,I2[m,d])
	for d in range(D):
		t5 = load(I2[m,d])
		t6 += t5
	t7 = 1/sqrt(t6)
	forall k in range(K):
		forall d in range(D):
			t8 = load(X[m,d])
			t9 = load(WT[k,d])
			t10 = dot(t8,t9)
			store(t10,I3[m,k,d])
		for d in range(D):
			t11 = load(I3[m,k,d])
			t12 += t11
		store(t12,I4[m,k])
	forall k in range(K):
		t13 = load(I4[m,k])
		t14 = row_scale(t13,t7)
		store(t14,I5[m,k])
	forall k in range(K):
		t15 = load(I5[m,k])
		t16 = t15/(1+exp(-t15))
		store(t16,I6[m,k])
	forall k in range(K):
		forall d in range(D):
			t17 = load(X[m,d])
			t18 = load(VT[k,d])
			t19 = dot(t17,t18)
			store(t19,I7[m,k,d])
		for d in range(D):
			t20 = load(I7[m,k,d])
			t21 += t20
		store(t21,I8[m,k])
	forall k in range(K):
		t22 = load(I8[m,k])
		t23 = row_scale(t22,t7)
		store(t23,I9[m,k])
	forall k in range(K):
		t24 = load(I6[m,k])
		t25 = load(I9[m,k])
		t26 = mul(t24,t25)
		store(t26,I10[m,k])
	forall n in range(N):
		forall k in range(K):
			t27 = load(I10[m,k])
			t28 = load(UT[n,k])
			t29 = dot(t27,t28)
			store(t29,I11[m,n,k])
		for k in range(K):
			t30 = load(I11[m,n,k])
			t31 += t30
		store(t31,O[m,n])
\end{lstlisting}
\end{minipage}

\paragraph{RMS+FNN-SwiGLU, Step 12: Fuse Map and Reduction} Still focused on the inner graph, the algorithm fuses the $D$-map surrounding the \texttt{row\_sum} operation with the subsequent reduction operation using \cref{rule:mapreduce}.\\ \\
\centerline{\includegraphics{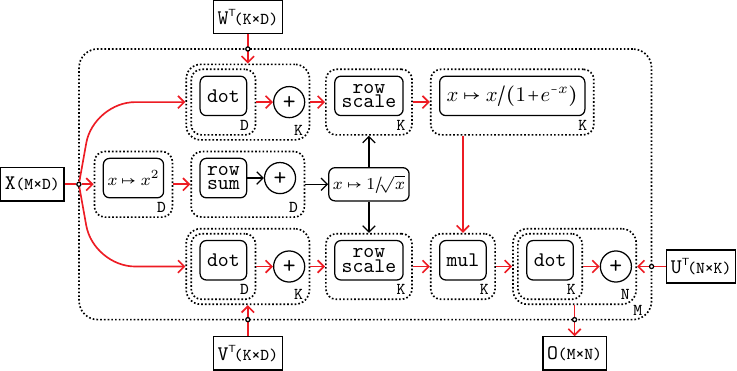}}

\begin{minipage}{\textwidth}
This block-program represented in code:
\begin{lstlisting}[multicols=2,numbers=left,xleftmargin=17pt]
forall m in range(M):
	forall d in range(D):
		t1 = load(X[m,d])
		t2 = t1**2
		store(t2,I1[m,d])
	for d in range(D):
		t3 = load(I1[m,d])
		t4 += row_sum(t3)
	t5 = 1/sqrt(t4)
	forall k in range(K):
		forall d in range(D):
			t6 = load(X[m,d])
			t7 = load(WT[k,d])
			t8 = dot(t6,t7)
			store(t8,I2[m,k,d])
		for d in range(D):
			t9 = load(I2[m,k,d])
			t10 += t9
		store(t10,I3[m,k])
	forall k in range(K):
		t11 = load(I3[m,k])
		t12 = row_scale(t11,t5)
		store(t12,I4[m,k])
	forall k in range(K):
		t13 = load(I4[m,k])
		t14 = t13/(1+exp(-t13))
		store(t14,I5[m,k])
	forall k in range(K):
		forall d in range(D):
			t15 = load(X[m,d])
			t16 = load(VT[k,d])
			t17 = dot(t15,t16)
			store(t17,I6[m,k,d])
		for d in range(D):
			t18 = load(I6[m,k,d])
			t19 += t18
		store(t19,I7[m,k])
	forall k in range(K):
		t20 = load(I7[m,k])
		t21 = row_scale(t20,t5)
		store(t21,I8[m,k])
	forall k in range(K):
		t22 = load(I5[m,k])
		t23 = load(I8[m,k])
		t24 = mul(t22,t23)
		store(t24,I9[m,k])
	forall n in range(N):
		forall k in range(K):
			t25 = load(I9[m,k])
			t26 = load(UT[n,k])
			t27 = dot(t25,t26)
			store(t27,I10[m,n,k])
		for k in range(K):
			t28 = load(I10[m,n,k])
			t29 += t28
		store(t29,O[m,n])
\end{lstlisting}
\end{minipage}

\paragraph{RMS+FNN-SwiGLU, Step 13-18: Fuse $K$-Maps and $D$-Maps} Still focused on the same graph, the algorithm applies \cref{rule:consecutive} or \cref{rule:sibling} five times to fuse all the $K$-maps, and once more to fuse two $D$-maps. \\ \\
\centerline{\includegraphics{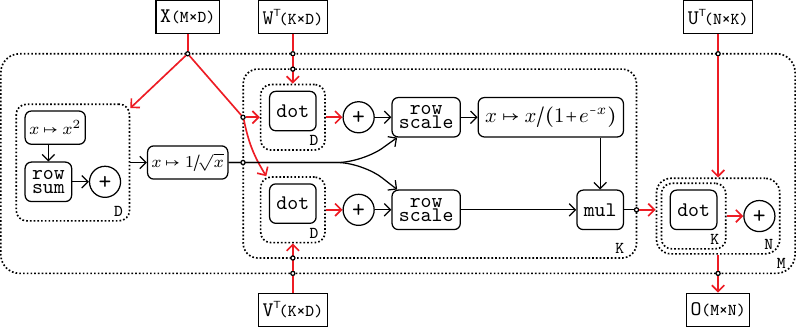}}

\begin{minipage}{\textwidth}
This block-program represented in code:
\begin{lstlisting}[multicols=2,numbers=left,xleftmargin=17pt]
forall m in range(M):
	for d in range(D):
		t1 = load(X[m,d])
		t2 = t1**2
		t3 += row_sum(t2)
	t4 = 1/sqrt(t3)

	forall k in range(K):
		forall d in range(D):
			t5 = load(X[m,d])
			t6 = load(WT[k,d])
			t7 = dot(t5,t6)
			store(t7,I1[m,k,d])
		for d in range(D):
			t8 = load(I1[m,k,d])
			t9 += t8
		t10 = row_scale(t9,t4)
		t11 = t10/(1+exp(-t10))
		forall d in range(D):
			t12 = load(X[m,d])
			t13 = load(VT[k,d])
			t14 = dot(t12,t13)
			store(t14,I2[m,k,d])
		for d in range(D):
			t15 = load(I2[m,k,d])
			t16 += t15
		t17 = row_scale(t16,t4)
		t18 = mul(t11,t17)
		store(t18,I3[m,k])
	forall n in range(N):
		forall k in range(K):
			t19 = load(I3[m,k])
			t20 = load(UT[n,k])
			t21 = dot(t19,t20)
			store(t21,I4[m,n,k])
		for k in range(K):
			t22 = load(I4[m,n,k])
			t23 += t22
		store(t23,O[m,n])
\end{lstlisting}
\end{minipage}

\paragraph{RMS+FNN-SwiGLU, Step 19-20: Fuse Maps and Reductions inside the $K$-Map} There is nothing more to do in the current graph, so the breadth-first search proceeds to the next level. Say that it goes into the inner graph in the $K$-map first (it could equally go into the $N$-map first). There, the algorithm identifies two separate opportunities to apply \cref{rule:mapreduce} and fuse maps with reductions. \\ \\
\centerline{\includegraphics{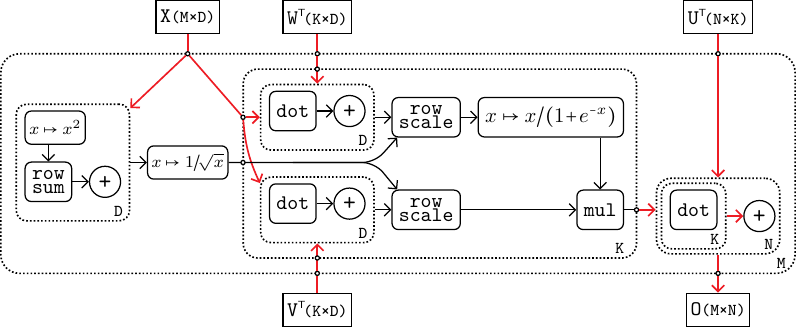}}

\begin{minipage}{\textwidth}
This block-program represented in code:
\begin{lstlisting}[multicols=2,numbers=left,xleftmargin=17pt]
forall m in range(M):
	for d in range(D):
		t1 = load(X[m,d])
		t2 = t1**2
		t3 += row_sum(t2)
	t4 = 1/sqrt(t3)
	forall k in range(K):
		for d in range(D):
			t5 = load(X[m,d])
			t6 = load(WT[k,d])
			t7 += dot(t5,t6)
		t8 = row_scale(t7,t4)
		t9 = t8/(1+exp(-t8))
		for d in range(D):
			t10 = load(X[m,d])
			t11 = load(VT[k,d])
			t12 += dot(t10,t11)
		t13 = row_scale(t12,t4)
		t14 = mul(t9,t13)
		store(t14,I1[m,k])
	forall n in range(N):
		for k in range(K):
			t15 = load(I1[m,k])
			t16 = load(UT[n,k])
			t17 = dot(t15,t16)
			store(t17,I2[m,n,k])
		for k in range(K):
			t18 = load(I2[m,n,k])
			t19 += t18
		store(t19,O[m,n])

\end{lstlisting}
\end{minipage}

\paragraph{RMS+FNN-SwiGLU, Step 21: Fuse Sibling $D$-Maps} Still inside the $K$-map, the algorithm fuses the two $D$-maps using \cref{rule:sibling}.\\ \\
\centerline{\includegraphics{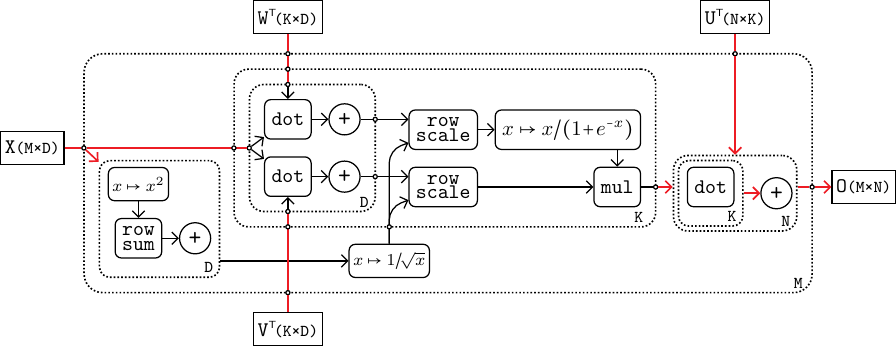}}

\begin{minipage}{\textwidth}
This block-program represented in code:
\begin{lstlisting}[multicols=2,numbers=left,xleftmargin=17pt]
forall m in range(M):
	for d in range(D):
		t1 = load(X[m,d])
		t2 = t1**2
		t3 += row_sum(t2)
	t4 = 1/sqrt(t3)
	forall k in range(K):
		for d in range(D):
			t5 = load(X[m,d])
			t6 = load(WT[k,d])
			t7 += dot(t5,t6)
			t8 = load(VT[k,d])
			t9 += dot(t5,t8)
		t10 = row_scale(t7,t4)
		t11 = t10/(1+exp(-t10))
		t12 = row_scale(t9,t4)
		t13 = mul(t11,t12)
		store(t13,I1[m,k])
	forall n in range(N):
		for k in range(K):
			t14 = load(I1[m,k])
			t15 = load(UT[n,k])
			t16 = dot(t14,t15)
			store(t16,I1[m,n,k])
		for k in range(K):
			t17 = load(I1[m,n,k])
			t18 += t17
		store(t18,O[m,n])
\end{lstlisting}
\end{minipage}

\paragraph{RMS+FNN-SwiGLU, Step 22: Fuse Maps and Reduction inside the $N$-Map} There is nothing more to do in the current graph, so the breadth-first search continues to the inner graph in the $N$-map, where the algorithm uses \cref{rule:mapreduce} to fuse a map with a reduction.\\ \\
\centerline{\includegraphics{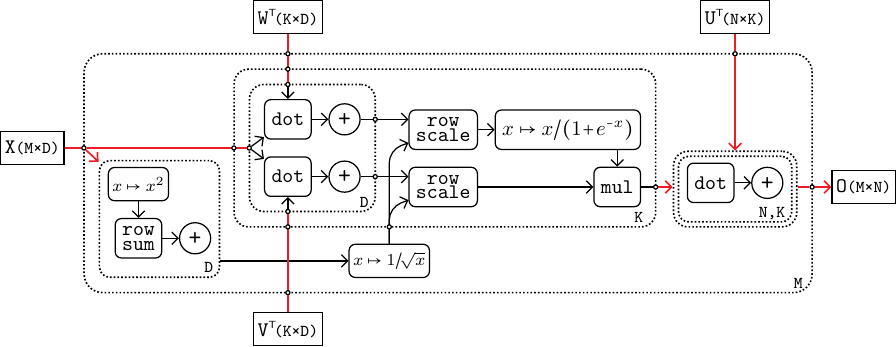}}

\begin{minipage}{\textwidth}
This block-program represented in code:
\begin{lstlisting}[multicols=2,numbers=left,xleftmargin=17pt]
forall m in range(M):
	for d in range(D):
		t1 = load(X[m,d])
		t2 = t1**2
		t3 += row_sum(t2)
	t4 = 1/sqrt(t3)
	forall k in range(K):
		for d in range(D):
			t5 = load(X[m,d])
			t6 = load(WT[k,d])
			t7 += dot(t5,t6)
			t8 = load(VT[k,d])
			t9 += dot(t5,t8)
		t10 = row_scale(t7,t4)
		t11 = t10/(1+exp(-t10))
		t12 = row_scale(t9,t4)
		t13 = mul(t11,t12)
		store(t13,I1[m,k])
	forall n in range(N):
		for k in range(K):
			t14 = load(I1[m,k])
			t15 = load(UT[n,k])
			t16 += dot(t14,t15)
		store(t16,O[m,n])
\end{lstlisting}
\end{minipage}

\paragraph{RMS+FNN-SwiGLU, Step 23: Extend $N$-Map} The \texttt{bfs\_fuse\_no\_extend} procedure has run out of matches at all levels of the block program. Therefore, the top-level procedure \texttt{fuse} captures a snapshot of the block-program and then calls \texttt{bfs\_extend}, which extends the $N$-map to the entire inner graph using \cref{rule:extend}. \\ \\
\centerline{\includegraphics{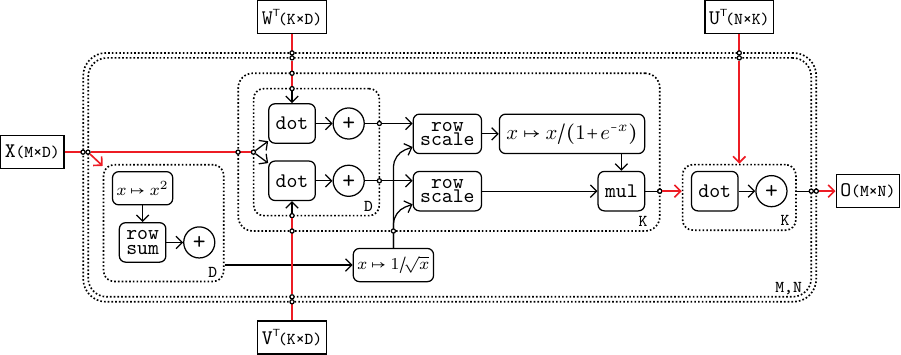}}

\begin{minipage}{\textwidth}
This block-program represented in code:
\begin{lstlisting}[multicols=2,numbers=left,xleftmargin=17pt]
forall m in range(M):
	forall n in range(N):
		for d in range(D):
			t1 = load(X[m,d])
			t2 = t1**2
			t3 += row_sum(t2)
		t4 = 1/sqrt(t3)
		forall k in range(K):
			for d in range(D):
				t5 = load(X[m,d])
				t6 = load(WT[k,d])
				t7 += dot(t5,t6)
				t8 = load(VT[k,d])
				t9 += dot(t5,t8)
			t10 = row_scale(t7,t4)
			t11 = t10/(1+exp(-t10))
			t12 = row_scale(t9,t4)
			t13 = mul(t11,t12)
			store(t13,I1[m,k])
		for k in range(K):
			t14 = load(I1[m,k])
			t15 = load(UT[n,k])
			t16 += dot(t14,t15)
		store(t16,O[m,n])
\end{lstlisting}
\end{minipage}

\paragraph{RMS+FNN-SwiGLU, Step 24: Fuse $K$-Maps} The breadth-first search starts from scratch and finds nothing to do in the top-level graph (which is a single $M$-map operator) or its inner graph (which is a single $N$-map operator). In the third level graph, it applies \cref{rule:consecutive} and fuses two $K$-maps.\\ \\
\centerline{\includegraphics{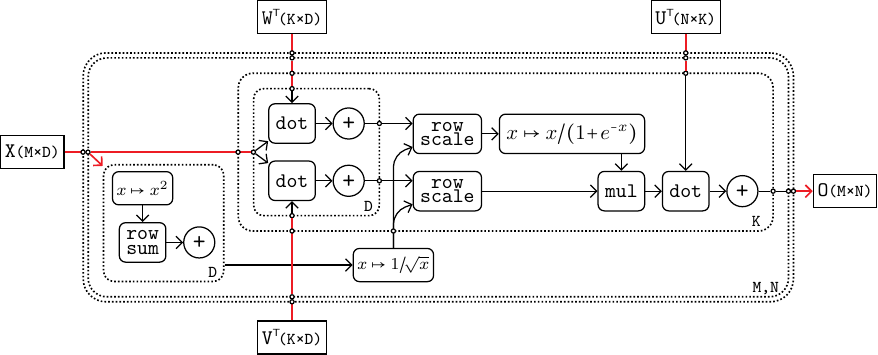}}

\begin{minipage}{\textwidth} 
This block-program represented in code:
\begin{lstlisting}[multicols=2,numbers=left,xleftmargin=17pt]
forall m in range(M):
	forall n in range(N):
		for d in range(D):
			t1 = load(X[m,d])
			t2 = t1**2
			t3 += row_sum(t2)
		t4 = 1/sqrt(t3)
		for k in range(K):
			for d in range(D):
				t5 = load(X[m,d])
				t6 = load(WT[k,d])
				t7 += dot(t5,t6)
				t8 = load(VT[k,d])
				t9 += dot(t5,t8)
			t10 = row_scale(t7,t4)
			t11 = t10/(1+exp(-t10))
			t12 = row_scale(t9,t4)
			t13 = mul(t11,t12)
			t14 = load(UT[n,k])
			t15 += dot(t13,t14)
		store(t15,O[m,n])
\end{lstlisting}
\end{minipage}

\paragraph{RMS+FNN-SwiGLU, Step 25: Extend $K$-Map} The \texttt{bfs\_fuse\_no\_extend} procedure has run out of matches at all levels of the block program, the top-level procedure \texttt{fuse} captures another snapshot of the block-program. It then calls \texttt{bfs\_extend}, which extends the $K$-map to the entire inner graph using \cref{rule:extend}.\\ \\
\centerline{\includegraphics{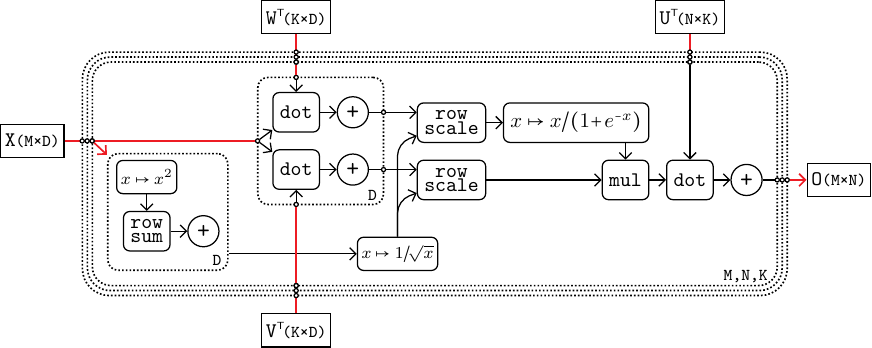}}

\begin{minipage}{\textwidth}
This block-program represented in code:
\begin{lstlisting}[multicols=2,numbers=left,xleftmargin=17pt]
forall m in range(M):
	forall n in range(N):
		for k in range(K):
			for d in range(D):
				t1 = load(X[m,d])
				t2 = t1**2
				t3 += row_sum(t2)
			t4 = 1/sqrt(t3)
			for d in range(D):
				t5 = load(X[m,d])
				t6 = load(WT[k,d])
				t7 += dot(t5,t6)
				t8 = load(VT[k,d])
				t9 += dot(t5,t8)
			t10 = row_scale(t7,t4)
			t11 = t10/(1+exp(-t10))
			t12 = row_scale(t9,t4)
			t13 = mul(t11,t12)
			t14 = load(UT[n,k])
			t15 += dot(t13,t14)
		store(t15,O[m,n])
\end{lstlisting}
\end{minipage}

\paragraph{RMS+FNN-SwiGLU, Step 26: Fuse $D$-Maps} The breadth-first search is started one last time, finds nothing to do in the first few levels of the block program, and eventually fuses two $D$-maps using \cref{rule:sibling}.\\ \\
\centerline{\includegraphics{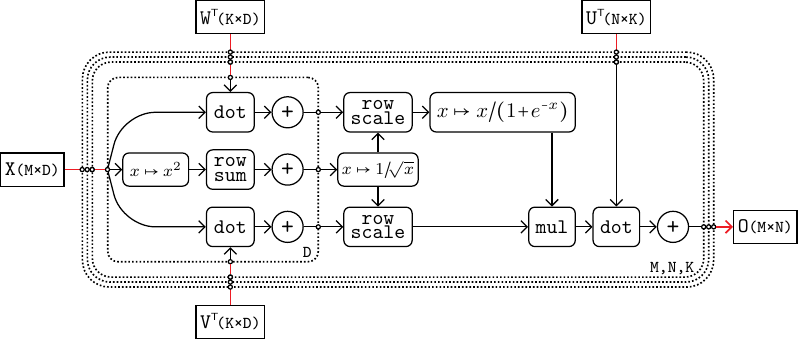}}

\begin{minipage}{\textwidth}
This block-program represented in code:
\begin{lstlisting}[multicols=2,numbers=left,xleftmargin=17pt]
forall m in range(M):
	forall n in range(N):
		for k in range(K):
			for d in range(D):
				t1 = load(X[m,d])
				t2 = t1**2
				t3 += row_sum(t2)
				t4 = load(WT[k,d])
				t5 += dot(t1,t4)
				t6 = load(VT[k,d])
				t7 += dot(t1,t6)
			t8 = 1/sqrt(t3)
			t9 = row_scale(t5,t8)
			t10 = t9/(1+exp(-t9))
			t11 = row_scale(t7,t8)
			t12 = mul(t10,t11)
			t13 = load(UT[n,k])
			t14 += dot(t12,t13)
		store(t14,O[m,n])
\end{lstlisting}
\end{minipage}

At this point, the fusion algorithm has exhausted all opportunities to match rules against any of the graphs.  

\paragraph{RMS+FNN-SwiGLU, Epilogue:} The only remaining buffered edges are those that are incident with input or output nodes, which means that the algorithm has successfully fused the entire program. The resulting program seems to do a lot of redundant work compared to the original program. However, recall that $N$ and $K$ are parameters that should be autotuned. In particular, the autotuner will consider setting either $N=1$, $K=1$, or both. If both $N=1$ and $K=1$, all the redundant work disappears, but the blocks may turn out to be too big to fit in local memory -- the autotuner could choose to compensate for this by increasing the values of $M$ and $D$. Alternatively, the autotuner could set only one of $N$ and $K$ to $1$ and set the other parameters in a way that balances redundant work versus block size.  

Another thing to notice is that the $K$-map in the final code is implemented as a serial \texttt{for} loop. Recall that this was an arbitrary choice made for the simplicity. This block program seems like an interesting candidate for a parallel reduction, for example one that is implemented with an atomic operation. For example, imagine setting $N=1$ and parallelizing both the $M$-map and the $K$-map across $MK$ parallel processors. 

As in the previous examples, additional fusion steps may be possible, but they depend on specific characteristics of the target hardware and exceed the scope of this paper. For example, on a CPU target, which computes individual block operators using loops, one could further fuse lines 6,7,9,11 and lines 12-18. 

\twocolumn

\section{Conclusion and Future Work}

We presented parts of the Blockbuster framework for operator fusion, which optimizes AI workloads running on any multiprocessor target with a tiered memory. We introduced the block program representation, which explicitly models how blocks of data move between memory tiers and presented a rule-based fusion algorithm that relies on this representation. While rule-based fusion is not a new idea, our algorithm achieves impressive results that exceed the current state-of-the-art. Its most notable result is the fused mega-kernel that we nicknamed Flash-RMSNorm+FFN-SwiGLU.

Our rule-based fusion algorithm relies on a small set of rules. One can easily think of additional logic-preserving substitutions that could be added to our framework (such as rules due to associative, distributive, and commutative laws for different algebraic operations), but we chose not to include them in the current version of our algorithm. Our choices were guided by the specific examples that we presented at the end of this paper. Going forward, as we explore new fusion examples, we will likely identify additional rules that are missing from our framework.  

The Blockbuster framework also includes a fusion-candidate selection algorithm, whose details are deferred to an upcoming companion paper \citep{Dekel2025b}. While the rule-based fusion algorithm is heuristic in nature, the candidate selection algorithm is a rigorous dynamic program, with strong optimality guarantees. It is disappointing that our rule-based fusion heuristic does not share similar theoretical guarantees, and we hope to address this in future work. 

\bibliographystyle{plainnat}
\bibliography{bib}
\clearpage

\appendix
\section*{Appendix: Adding Numerical Safety}
The main part of this paper ignored numerical safety by assuming that arithmetic operations are conducted with infinite-precision real numbers. In practice, modern computers use floating-point arithmetic, which faces the risk of numerical overflow, especially when dealing with the exponents. Luckily, AI compilers can identify all exponential operations and make them numerical safe using a separate compiler pass, which comes after all the fusion passes.  

The key to numerical safety is to represent exponentiated numbers using two variables, a \emph{significant} and an \emph{exponent}. In other words, we represent the real number $x$ as $s e^t$, where $(s,t)$ is a pair of floating-point variables called the significand-exponent pair. This is essentially a software floating point implementation which is constructed on top of the hardware floating point representation. 

Any number number $x$ can be easily represented as the significand-exponent pair $(x,0)$. Any exponent $e^y$ can be represented as $(1,y)$. The inverse of $(s,t)$ is $(1/s, -t)$. Multiplying two numbers, $(s_1,t_1)$ and $(s_2,t_2)$, results in 
$$
(s_1,t_1) \cdot (s_2,t_2) ~=~
(s_1 s_2, t_1+t_2).
$$
Adding two numbers $(s_1,t_1)$ and $(s_2,t_2)$ requires an additional variable $z$, whose value will be determined shortly. Note that the following identify holds for any value of $z$:
$$
(s_1,t_1) + (s_2,t_2) ~=~ (s_1 e^{t_1-z} + s_2 e^{t_2-z}, ~z).
$$
To ensure that $e^{t_1-z}$ and $e^{t_2-z}$ do not overflow, we can choose $z = \max(t_1,t_2)$, ensuring that both values are between $0$ and $1$. If $s_1, s_2, t_1, t_2$ are stored as integers or low-precision types (such as FP8 or FP6), a smaller value of $z$ may do a better job of balancing the risks of numerical overflow and underflow. A compiler can identify all the variables that face numerical risks and replace their representation with a significand-exponent pair.

We can further develop this idea from individual numbers to entire matrix blocks. Each element in the block will have a separate significand, but they will all share a common exponent. We write the pair as $(S,t)$, where $S$ is a block of significands and $t$ is their shared exponent. A standard block of floating point values $X$ is represented as $(X,0)$ and the element-wise exponent $e^X$ is represented as $(e^{X-z},z)$ where $z = \max(X)$ (or a smaller value, which balances the risks numerical overflow and underflow). As before, addition is defined as
$$
(S_1,t_1) + (S_2,t_2) ~=~ (S_1 e^{t_1-z} + S_2 e^{t_2-z}, ~z),
$$
with $z = \max(t_1,t_2)$ (or a smaller value, which balances the risks numerical overflow and underflow). Matrix multiplication of the two blocks is defined as
$$
(S_1,t_1) \cdot (S_2,t_2) ~=~ (S_1 \cdot S_2, t_1+t_2).
$$

Sharing an exponent across an entire block is cheaper than keeping an exponent per element, but runs the risk of underflowing very small values. An intermediate approach is to keep a separate exponent for each row in the block, which is the approached used in the Flash Attention paper (referred to as \emph{online softmax}, but really has nothing specific to SoftMax). All of these approaches (separate exponents, row-wise shared exponent, block shared exponent) are equally safe, and vary only in their computational cost and their precision. The compiler should consider each variant and apply the one that optimally trades-off cost and precision. 

\end{document}